\documentclass{ucetd}
\usepackage{subfigure,epsfig,amsfonts}
\usepackage{amsmath}
\usepackage{amssymb}
\usepackage{amsthm}

\usepackage[utf8]{inputenc} 
\usepackage[T1]{fontenc}    
\usepackage{url}            
\usepackage{booktabs}       
\usepackage{amsfonts}       
\usepackage{nicefrac}       
\usepackage{microtype}      

\usepackage[textsize=scriptsize]{todonotes} 
\usepackage{lineno}

\usepackage{color}
\usepackage{xspace}
\makeatletter
\usepackage{tikz}
\usetikzlibrary{calc}
\newcommand*\circled[1]{\tikz[baseline=(char.base)]{
    \node[shape=circle, draw, inner sep=1pt,
        minimum height={\f@size*1.6},] (char) {\vphantom{WAH1g}#1};}}
\makeatother

\makeatletter
\renewcommand{\todo}[2][]{%
    \@todo[caption={#2}, #1]{\begin{spacing}{0.5}#2\end{spacing}}%
}
\makeatother

\usepackage{graphicx, amsmath, amssymb, amsthm, bbm, mathtools, nicefrac, algorithm, algorithmic, float, txfonts, centernot, xspace, adjustbox}
\usepackage[style=alphabetic,natbib=true,maxbibnames=99]{biblatex}
\theoremstyle{definition}
\newtheorem{assum}{Assumption}[section]
\newtheorem{defn}{Definition}[section]

\theoremstyle{plain}
\newtheorem{thm}{Theorem}[section]
\newtheorem{lem}[thm]{Lemma}
\newtheorem{pro}[thm]{Proposition}
\newtheorem{cor}[thm]{Corollary}

\theoremstyle{remark}
\newtheorem*{rem}{Remark}

\usepackage{tikz}
\usetikzlibrary{arrows, automata, positioning}

\newcommand{\Prob}{\mathbb{P}}
\newcommand{\Exp}{\mathbb{E}}
\newcommand{\Ind}{\mathbbm{1}}
\newcommand{\nPerp}{\centernot{\Perp}}

\DeclarePairedDelimiter{\floor}{\lfloor}{\rfloor}
\DeclarePairedDelimiter{\sbk}{[}{]}
\DeclarePairedDelimiter{\rbk}{(}{)}

\DeclarePairedDelimiter{\cbk}{\{}{\}}
\DeclarePairedDelimiter{\vbk}{|}{|}
\DeclarePairedDelimiter{\dvbk}{\|}{\|}

\newcommand{\rmax}{r_\text{max}\xspace}
\newcommand{\states}{\mathcal{S}\xspace}
\newcommand{\actions}{\mathcal{A}\xspace}
\newcommand{\reset}{\texttt{RESET}\xspace}
\newcommand{\ucrl}{\texttt{UCRL2}\xspace}

\newcommand\numberthis{\addtocounter{equation}{1}\tag{\theequation}}

\title{On Reward Structures \\of Markov Decision Processes}
\author{Falcon Z.~Dai}
\department{Thesis Department}
\division{Thesis Division}
\degree{Doctor of philosophy}
\date{September, 2022}

\epigraph{Live as if you were to die tomorrow. Learn as if you were to live forever. \\---Mahatma Ghandi}

\usepackage[pdfusetitle]{hyperref}
\hypersetup{unicode=true,
            linktoc=all,
            pdfsubject=subject here,
            pdfkeywords=keyword1 keyword2 keyword3,
            pdfborder={0 0 0},
            breaklinks=true}
\makeatletter
\let\ORG@hyper@linkstart\hyper@linkstart
\protected\def\hyper@linkstart#1#2{%
  \lowercase{\ORG@hyper@linkstart{#1}{#2}}}
\makeatother

\begin{document}
\maketitle

\makeepigraph

\tableofcontents
\listoffigures

\acknowledgments


My journey into the world of machine learning started in the summer of 2012 after my graduation from the University of Chicago. I was fascinated by the possibility of a learning machine, not unlike what Alan Turing had envisioned as a path to artificial intelligence \cite{turing1950computing}. A decade later, I am delighted to have acquired the modern language to frame the obstacles to that dream and made my small contributions to that endeavor. Along the way, I experienced the joy of unexpected discovery and the excitement of recognition. I remained frequently amazed by how the study of reinforcement learning could provide ideas for living a better life.

First and foremost, I would like to thank my advisor, Matt Walter. His support and guidance have enabled my exploration of many ideas some of which are presented here. His work ethics and his kindness will always be an inspiration to me.

There are many people whose company I am deeply thankful for. Without them, my intellectual voyage would have been unbearably lonely: Zheng Cai, Andrea Daniele, Chip Schaff, David Yunis, Lifu Tu, Shubham Toshniwal, Jiading Feng, Igor Vasiljević, Ruotian Luo, Takuma Yoneda, Shengjie Lin, Jon Michaux, Daping Weng, Payman Yadollahpour, Ruixi Mao, Ronan Fruit, Audrey Sedal, Karl Stratos, Kevin Gimpel, Greg Shakhnarovich, Leo Towle, David McAllester, Sasha Rush, Avrim Blum, Julia Chuzhoy, Mary Silber and many others.

With the generous support of TTIC, my many intellectual interests have gone further than I imagined \cite{dai2017glyph,dai2019towards}. Mary Marre, Amy Minick, Jessica Jacobson, Adam Bohlander has made me feel part of a family.

Lastly, I wish to thank my parents for their patience and love: You Dai, whose liking of philosophy has shaped my taste; Xuetian Wu, whose resilience taught me not to give up. During the last years of my study, I have been fortunate enough to form my own family: Yunxi Duan, whose pragmatism and grace have greatly enriched my life.

\abstract

A Markov decision process can be parameterized by a transition kernel and a reward function.
Both play essential roles in the study of reinforcement learning as evidenced by their presence in the Bellman equations.
In our inquiry of various kinds of ``costs'' associated with reinforcement learning inspired by the demands in robotic applications, rewards are central to understanding the structure of a Markov decision process and reward-centric notions can elucidate important concepts in reinforcement learning.

Specifically, we study the sample complexity of policy evaluation and develop a novel estimator with an instance-specific error bound of $\widetilde{O}(\sqrt{\nicefrac{\tau_s}{n}})$ for estimating a single state value. Under the online regret minimization setting, we refine the transition-based MDP constant, diameter, into a reward-based constant, maximum expected hitting cost, and with it, provide a theoretical explanation for how a well-known technique, potential-based reward shaping, could accelerate learning with expert knowledge. In an attempt to study safe reinforcement learning, we model hazardous environments with irrecoverability and proposed a quantitative notion of safe learning via reset efficiency. In this setting, we modify a classic algorithm to account for resets achieving promising preliminary numerical results. Lastly, for MDPs with multiple reward functions, we develop a planning algorithm that computationally efficiently finds Pareto-optimal stochastic policies.

\mainmatter

\chapter{Introduction}
Reinforcement learning (RL) is concerned with sequential decision making in an unknown environment. The environment is modeled as a discrete-time finite Markov decision process (MDP) in the context of this thesis as in classical literature. We interact with an MDP by choosing an action (based on history) and this decision affects transition probabilities over the next states. We receive an immediate reward based on the current state and the action we choose. The sum of immediate rewards would be the primary rubric for measuring the merits of a policy or an algorithm that we follow. Despite the initial ignorance of the actual environment, it is expected that through properly chosen actions, we can learn more and more about the environment and gradually improve the rewards we obtain and the policy we follow. Thanks to the flexibility of MDPs, RL has been applied to many domains and demonstrated superior performance, including games \cite{silver2016mastering,mnih2015human}, language modeling \cite{ouyang2022training}, robotics \cite{kober2013reinforcement} among others.

Being surrounded by colleagues that are concerned with the applicability of RL in real-world robotics (in developing optimal controllers), we are inspired to formalize some of the challenges and analyze them. Due to the nature of reinforcement learning, it generally takes many interactions to learn a good policy. Thus we study the ``cost'' of reinforcement learning and ways to lower it.
The main concerns that guide our investigations revolve around the ``cost'' of learning are:
\begin{itemize}
  \item How many samples (interactions) do we need to learn?
  \item Can we learn faster with some extra help from an expert?
  \item Can we learn safely incurring few costly mistakes?
\end{itemize}

After introducing some prerequisite definitions and notational conventions, we shall present the various investigations roughly in an ascending order of complexities and generality of their assumptions.
In Chapter~\ref{ch:policy-evaluation}, we consider the policy evaluation problem where we wish to evaluate how rewarding a given policy is via samples. We present a novel policy evaluation routine called \emph{loop estimator} that enables efficiently evaluating the value of a single state, a capability absent in other methods~\cite{dai2021loop}.
In Chapter~\ref{ch:expert-knowledge}, we present a novel analysis of a well-known practical method of providing expert knowledge to an RL algorithm, \emph{potential-based reward shaping}~\cite{ng1999policy}. Despite being a popular method in applied RL studies, reward shaping is not well understood formally. By constructing a new MDP constant, \emph{maximum expected hitting cost} (MEHC), that measures how costly it is to move from one state to another within an MDP, we are able to show that reward shaping changes the maximum expected hitting cost of an MDP, and thereby it changes the ``speed'' of learning~\cite{dai2019maximum}. Surprisingly, we also discover that the effect of reward shaping on MEHC is limited to a multiplicative factor of two~\cite{dai2019maximum}.
In Chapter~\ref{ch:safety}, we examine the problem of safety in the context of RL. Firstly, we discuss a more general class of MDPs, \emph{multichain} MDPs, and show that they can model environments containing irrecoverable actions. Secondly, we recast resets as actions (and not an automatic or involuntary transition as in episodic MDPs) and we shall evaluate how safe a learning algorithm quantitatively is by counting the resets it takes. Lastly, by framing reset counts as a second sort of reward in an environment, we consider the problem of finding Pareto-optimal policies in environments with multiple reward functions.
In Chapter~\ref{ch:conclusion}, we conclude with thoughts on future directions.


\section{Markov decision processes}

We model our interactive environments with (discrete-time) Markov decision processes, i.e., the stochastic processes are indexed by $t \in \mathbb{N}$.

\begin{defn}[discrete-time Markov decision process]
  Given
  a state space $\mathcal{S}$,
  an action space $\mathcal{A}$,
  an action sequence $(A_t)_{t \ge 0}$ where $A_t \in \actions$,
  a transition kernel $p : \mathcal{S} \times \mathcal{A} \rightarrow \mathcal{P}\rbk*{\mathcal{S}}$,
  and rewards $r : \mathcal{S} \times \mathcal{A} \rightarrow \mathcal{P}\rbk*{\sbk*{0, \rmax}}$,
  a (discrete-time) \emph{Markov decision process} (MDP) $\rbk*{S_t, A_t, R_t}_{t \ge 0}$ is a stochastic process
  in which states transition according to $p$, i.e., $S_{t+1} \sim p(S_t, A_t)$
  and rewards are drawn independently and identically from $r$, i.e., $R_t \sim r(S_t, A_t)$.
\end{defn}

A \emph{finite} MDP has a finite state space and a finite action space, i.e., $S \coloneqq \vbk*{\mathcal{S}} < \infty$ and \mbox{$A \coloneqq \vbk*{\mathcal{A}} < \infty$}.
It is often sufficient to characterize the $\sbk*{0, \rmax}$-bounded reward distributions up to their means denoted by \mbox{$\bar{r}(s, a) \coloneqq \Exp\sbk*{R_0 | S_0=s, A_0=a}$}.

As suggested in the name Markov \emph{decision} process, we need to decide the sequence of actions $\rbk*{A_t}_{t \ge 0}$ and thereby affect the states and rewards. We may follow a policy to choose the actions. Formally, a (deterministic and Markov) \emph{policy} $\pi : \mathcal{S} \rightarrow \mathcal{A}$ recommends what actions to take at each state, akin to a feedback controller.
Relatedly, a stochastic Markov policy chooses an action based on the state according to a distribution over actions, $\mathcal{S} \rightarrow \mathcal{P}\rbk*{\mathcal{A}}$. Since the smaller class of deterministic policies achieve the same optimal long-term rewards as the strictly larger class in many settings \cite{puterman1994markov}, we will mainly study the deterministic policies and revisit the stochastic policies in Section~\ref{sec:multiple-rewards}.
Similarly, an \emph{algorithm} $\mathfrak{L}$ at step $n$ can recommend an action based on the process $(S_t, A_t, R_t)_{0 \le t < n}$.

In reinforcement learning, we typically consider $p$ and $\bar{r}$ to be unknown to the algorithm and $\mathcal{S}$, $\mathcal{A}$, and $\rmax$ to be given. We may refer to an MDP by its parameters $\mathcal{M} \coloneqq (\mathcal{S}, \mathcal{A}, p, r)$.
We will abbreviate $\Exp_s\sbk*{\cdot} \coloneqq \Exp\sbk*{\cdot \vert S_0 = s}$
and $\Exp_s^\pi\sbk*{\cdot} \coloneqq \Exp\sbk*{\cdot \vert S_0 = s, A_t=\pi(S_t)}$.

For finite MDPs, we can conveniently represent the transition kernel $p$ as a rank-3 tensor $\mathbf{P}$ with index from $[S] \times [A] \times [S]$, where $[n] = \cbk*{1, \cdots, n}$ such that
$$ P_{s, a, s'} \coloneqq \Prob\sbk*{S_1 = s' \vert S_0 = s, A_0 = a}. $$
Similarly, when given a policy $\pi$, we can present the transition kernel as a row-stochastic matrix $\mathbf{P}^\pi$, where
$$ P^\pi_{s, s'} \coloneqq \Prob\sbk*{S_1 = s' \vert S_0 = s, A_0 = \pi(s)}. $$
Furthermore, distributions over states are represented by row vectors, whereas functions over states, column vectors.

\begin{rem}
  MDPs are very flexible models as they allow for stochastic transitions, unlike the deterministic dynamics typically studied in (optimal) control theory.
\end{rem}

\section{Related processes}

There are several notable stochastic processes related to MDPs.
\begin{defn}[Markov reward process]
  Given
  a state space $\mathcal{S}$,
  a transition kernel $p : \mathcal{S} \rightarrow \mathcal{P}\rbk*{\mathcal{S}}$,
  and rewards $r : \mathcal{S} \rightarrow \mathcal{P}\rbk*{\sbk*{0, \rmax}}$,
  a \emph{Markov reward process} (MRP) $\rbk*{S_t, R_t}_{t \ge 0}$ is a stochastic process
  in which states transition according to $p$, i.e., $S_{t+1} \sim p(S_t)$
  and rewards are drawn independently and identically from $r$, i.e., $R_t \sim r(S_t)$.
\end{defn}

Given an MDP and a policy $\pi : \mathcal{S} \rightarrow \mathcal{A}$, if we take actions according to $\pi$, i.e., $A_t = \pi(S_t)$, then states transition according to $S_{t+1} \sim p(S_t, \pi(S_t))$ and rewards distribute according to \mbox{$R_t \sim r(S_t, \pi(S_t))$}. Therefore $\rbk*{S_t, R_t}$ is an MRP of transition kernel $p(\cdot, \pi(\cdot))$ and rewards $r(\cdot, \pi(\cdot))$.
%
\begin{defn}[Markov chains]
  Given
  a state space $\mathcal{S}$,
  and a transition kernel $p : \mathcal{S} \rightarrow \mathcal{P}\rbk*{\mathcal{S}}$,
  a \emph{Markov chain} (MC) $\rbk*{S_t}_{t \ge 0}$ is a stochastic process
  in which states transition according to $p$, i.e., $S_{t+1} \sim p(S_t)$.
\end{defn}

Observe that $\rbk*{S_t}_{t \ge 0}$ is an MC embedded in an MRP $\rbk*{S_t, R_t}_{t \ge 0}$ when we omit the rewards.

The key property that MDP, MRP and MC share is the Markov property. Informally speaking, it means that the ``memory'' of a process is limited to the \emph{latest} state (or state-action in the case of an MDP) with respect to a linear order of the index. This linearly ordered index is often thought of as time. For example, in a Markov chain, we have
$$ \Prob\sbk*{S_{t+1} \vert S_0, \cdots , S_t} = \Prob\sbk*{S_{t+1} \vert S_t}. $$
In contrast, a set of independently and identically distributed (IID) random variables have no ``memory.'' One can also extend the single step ``memory'' to $n$-many steps and this is called $n$-step Markov property. In online learning, one often considers a process that has unlimited memory and is adversarial.

One important phenomenon in finite Markov chains is the existence of stationary distributions. Using the matrix notations, we call $\sigma \in \mathcal{P}\rbk*{\sbk*{S}}$ a \emph{stationary distribution} of a Markov chain with transition kernel $\mathbf{P}$ if
$$ \sigma \mathbf{P} = \sigma. $$
Therefore, $\sigma$ is a left eigenvector of $\mathbf{P}$ corresponding to the eigenvalue of one. $\mathbf{P}$ must have an eigenvalue of one because it is a stochastic matrix where each row sums to one. One way to construct the stationary distribution $\sigma$ is via the limit of the running distribution of states
$$ \lim_{n\rightarrow\infty} \frac{1}{n} \sum_{i=1}^n x \mathbf{P}^i, $$
where $x \in \mathcal{P}\rbk*{\sbk*{S}}$ is an arbitrary distribution over $[S]$.
If $\mathbf{P}$ is aperiodic, then a faster converging limit can be used, c.f., power iteration algorithm,
$$ \lim_{n\rightarrow\infty} x \mathbf{P}^n. $$
The stationary distribution of a Markov chain need not be unique.

We can extend discrete-time MDPs slightly to continuous time while retaining the discrete state transitions by allowing the transitions to occur at discrete random times.
\begin{defn}[semi-Markov decision process]
  Given
  a state space $\mathcal{S}$,
  an action space $\mathcal{A}$,
  an action sequence $(A_t)_{t \ge 0}$ where $A_t \in \actions$,
  a transition kernel $p : \mathcal{S} \times \mathcal{A} \rightarrow \mathcal{P}\rbk*{\mathcal{S} \times \rbk*{0, \infty}}$,
  and reward rates $r : \mathcal{S} \times \mathcal{A} \rightarrow \mathcal{P}\rbk*{\sbk*{0, \rmax}}$,
  a \emph{semi-Markov decision process} (SMDP) $\rbk*{S_t, A_t, R_t, I_t}_{t \ge 0}$ is a stochastic process
  in which state transitions and interarrival times distribute according to $p$, i.e., $S_{t+1}, I_t \sim p(S_t, A_t)$
  and reward rates are drawn independently and identically from $r$, i.e., $R_t \sim r(S_t, A_t)$ and the reward received over the transition would be $R_t I_t$.
\end{defn}
Given an MDP, if we think of committing to a policy until some termination condition is met and we decide on the next action, then the transitions occur at random times and we can model the resulting process as an SMDP \cite{sutton1999between}. In this case, the random waiting times till transitions are integers.

If we remove states from MDPs instead of actions (as we did in MRPs), then we get an important model called multi-armed bandits (MAB). MABs are widely studied in online learning literature as they encapsulate many problems that require some ``exploration vs exploitation'' tradeoff. Similar to MDPs, we need to make decisions as the action sequence is left unspecified.
\begin{defn}[Multi-armed bandits]
  Given
  an action space $\actions$,
  an action sequence $(A_t)_{t \ge 0}$ where $A_t \in \actions$,
  and rewards $r : \actions \rightarrow \mathcal{P}\rbk*{\sbk*{0, \rmax}}$,
  a \emph{multi-armed bandit} (MAB) $\rbk*{A_t, R_t}_{t \ge 0}$ is a stochastic process for which $R_t \sim r(A_t)$.
\end{defn}

A particularly nice algorithm \texttt{UCB} (upper confidence bound) \cite{auer2002finite} for solving RL problems in MABs provides good intuition known as ``optimism in the face of uncertainty'' (OFU) and has been extended to \texttt{UCRL} (upper confidence bound for reinforcement learning), which solves RL problems in MDPs \cite{jaksch2010near}. We will follow this general idea in learning. A popular alternative is Thompson sampling which presents some computational difficulty \cite{fruit2018near}.

%

\section{Values}

We shall evaluate the merit of a policy by the expected long-term rewards it receives. By ``long-term,'' we mean to distinguish the cumulative rewards over many time steps from the immediate rewards that we receive at each step.
More concretely, we consider the rewards we receive over infinitely many time steps, c.f., summing over a specific finite time horizon.
We use two common evaluation criteria that incorporate immediate rewards over an infinite time horizon: exponentially discounted value and average reward.

\begin{defn}[Discounted value]
  Given an MRP $\rbk*{S_t, R_t}_{t \ge 0}$, \emph{discount} $\gamma \in [0, 1)$, and a state $s \in \mathcal{S}$, we define \emph{the ($\gamma$-discounted) value of state $s$} as
  \begin{equation}\label{eq:discounted-value}
    v(s) \coloneqq \Exp_s\sbk*{\sum_{t=0}^\infty \gamma^t R_t }.
  \end{equation}
\end{defn}

This quantity is well-defined since $R_t \in [0, \rmax]$. In the case of an MDP with a stationary and Markov policy $\pi$, we write $ v^\pi(s) \coloneqq \Exp_s^\pi\sbk*{\sum_{t=0}^\infty \gamma^t R_t } $ and the optimal value $ v^*(s) \coloneqq \max_{\pi : \mathcal{S} \rightarrow \mathcal{A}} v^\pi(s)$. The choice of exponential discounting might seem arbitrary at first. Two equivalent (in expectation) models using undiscounted total rewards can provide some intuitions. One is to imagine that there is an interest accruing at a fixed rate of $(1 - \gamma)$ on the rewards obtained and that we report the total earnings in terms of the initial value of money. The other is to imagine that the process may terminate with a fixed probability of $(1 - \gamma)$ at the current step and all further rewards after termination are zero.

There exists a policy (not necessarily unique) that achieves the optimal value of every state. This can be seen by considering the Bellman optimality equation whose solution, the optimal state value function, is unique.

%


One of the key early insights in the study of MDPs is the subproblem structure, e.g., dynamic programming. In the case of an infinite horizon discounted values, we have
\begin{equation}\label{eq:bellman-state}
  v^\pi(s) = \bar{r}^\pi(s) + \gamma \sum_{s'\in\states} P^\pi_{s, s'} v^\pi(s')
\end{equation}
where $\bar{r}^\pi(s) \coloneqq \Exp\sbk*{R_1 \vert S_1=s, A_1=\pi(S_1)}$ is the expected immediate reward at state $s$ following policy $\pi$.

In particular, the optimal value satisfies the optimal Bellman equations
\begin{equation}\label{eq:optimal-bellman-state}
  v^*(s) = \max_a \bar{r}(s, a) + \gamma \sum_{s'\in\states} P_{s, a, s'} v^*(s').
\end{equation}

\begin{defn}[Gain]\label{def:gain}
  Given an MDP with parameters $\mu = \rbk*{\mathcal{S}, \mathcal{A}, p, r}$ with a policy $\pi$, and a state $s \in \mathcal{S}$,
  the expected $n$-step cumulative reward starting in state $s$ is
  $$ c(\mu, \pi, n, s) \coloneqq \Exp^\pi_s \sbk*{\sum_{t=1}^n R_t} $$
  and the \emph{gain} (or \emph{average reward}) of a state $s$ is
  $$ g(\mu, \pi, s) \coloneqq \lim_{n\rightarrow\infty} \frac{1}{n} c(\mu, \pi, n, s) = \lim_{n\rightarrow\infty} \frac{1}{n} \Exp^\pi_s \sbk*{\sum_{t=1}^n R_t} $$
  and the \emph{optimal gain of state $s$}
  $$g^*(\mu, s) \coloneqq \max_{\pi \in \mathcal{S} \rightarrow \mathcal{A}} g(\mu, \pi, s).$$
\end{defn}

The (optimal) gain of a state is well-defined due to the existence of stationary distributions in a Markov chain and sometimes it is referred to as the stationary reward.

One nice consequence of using these criteria in the infinite horizon setting is that there exists value-maximizing \emph{stationary} policies \cite{puterman1994markov}. In contrast, there may not be value-maximizing stationary policies in the finite horizon setting, that is, the optimal actions might depend on both the current state and the time left till termination.

The analog of Bellman equations for gains is more complicated. For the states $\states$ that have the same gain $g$ under policy $\pi$, the gain $g$ and \emph{bias} $b : \states \rightarrow \mathbb{R}$ satisfy a set of equations,
\begin{equation}\label{eq:bias-gain}
  b(s) = \bar{r}^\pi(s) + \sum_{s'} P^\pi_{s, s'} b(s') - g.
\end{equation}

Biases measure the excess expected total rewards starting in a given state than starting in some reference distribution of states and they are only defined up to an additive constant \cite{puterman1994markov}. This constant reflects the choice of reference distribution.

In contrast to gains, which are determined by the rewards in the infinitely long tail when visited states converge to the stationary distribution, the discounted values mix the early rewards and the later ones while heavily favoring former. The two forms of values are related by the Laurent series expansion of the discounted value around the perturbation of the discount factor from one \cite{puterman1994markov}.

\section{Transition structures}

One way to classify states is based on what states can be reached from what other states. Recall that similar classification exists for MCs.

\begin{defn}[reachability]
  In an MC, we say that state $s'$ is \emph{reachable} from state $s$ if there exists $n \in \mathbb{N}$ such that $\Prob\sbk*{S_n=s'\vert S_1=s} > 0$.
\end{defn}

If we consider whether a state can appear infinitely many times in an MC, we can classify states as ergodic or transient. Furthermore, we can partition the states based on how they recur with other states.
A set of states are considered \emph{closed} if no other states are reachable from these states. If no proper subset of states is closed, then it is considered \emph{irreducible}. We call these sets of states closed \emph{irreducible recurrent classes}, or \emph{subchains} for short.
In general, an MC can have multiple subchains and the rest are called transient states. The transit states have finite expected visits.

By applying the the Fox-Landi algorithm~\cite{fox1968algorithm} to the transition matrix of an MC, we can find the partition of the states into subchains and a (possibly empty) set of transient states.

For an MDP, different policies can induce different subchains. We can nevertheless extend many notions in MC to MDP by ranging over all policies.
State $s$ \emph{communicates} with state $s'$ if there exist policies $\pi$ such that $s$ and $s'$ both belong to the same subchain.

If all states communicate with all other states, then the MDP is a \emph{communicating} MDP. One important consequence of communication is that the \emph{optimal} gain is the same for all states.

We can quantify how ``large'' a communicating MDP is by measuring how long it takes to visit any state from any other state. Let $H_s \coloneqq \inf \{n : S_n = s \}$ be the hitting time of state $s$, then
\begin{defn}[Diameter \cite{jaksch2010near}]
  Given an MDP $\mathcal{M}$, we define \emph{the diameter of $\mathcal{M}$} as
  \begin{equation}
    \tau(\mathcal{M}) \coloneqq \max_{s, s' \in \mathcal{S}} \min_{\pi \in \mathcal{S}\rightarrow\mathcal{A}} \Exp^\pi_s\sbk*{H_{s'}}.
  \end{equation}
\end{defn}

More generally, an MDP can have more than one subchain and can have states that are transient under all policies. This is called a \emph{multichain} MDP. For mulitchain MDPs, we can partition its state-action space so that each component forms a communicating MDP, that is, communicating when restricted to some actions, and the transient states.
Unique chain decomposition \cite{schweitzer1984value} is an algorithm that can find such subchain structures of a multichain MDP.
Once the MDP is decomposed into communicating subchains, for some computational purposes, we can view the original MDP as an SMDP over the subchains. Such a hierarchical representation is the basis for extending value iteration to multichain MDPs \cite{ohno1988value}.

\section{Problem formulations}
If we assume full knowledge of both the transition and reward functions, and wish to find an optimal policy. This is known as a planning problem and the main challenge is computational.
Near-optimal policies with respect to discounted value or gain can be obtained with efficient algorithms such as value iteration \cite{puterman1994markov}. These approximate solutions to planning are often relied upon to find good actions to take in the context of RL when we do not have full knowledge of the environment.

In reinforcement learning-theoretic settings, we typically assume little knowledge of the transition and reward functions of the actual environment. Under this weaker assumption, knowledge of the environment is gradually obtained through interactions thus presenting an additional challenge of statistical efficiency.

\subsection{Policy evaluation}
Without prior knowledge of the transitions or rewards, we are interested in evaluating the value of a given policy through samples. More formally, we wish to find algorithms that can accurately estimate the value of a policy $\pi$ given a truncated sample path of an MDP $\rbk*{S_t, A_t, R_t}_{1 \le t \le n}$, where $A_t = \pi(S_t)$. This sample path is considered ``on-policy'' as the actions are taken according to the policy we wish to evaluate. In a policy iteration algorithm, this estimation process interleaves with policy improvement in which we update the policy to be greedy with respect to the estimated values.
In contrast, one can imagine that the sample path is collected with a policy different other than $\pi$ and that will be considered ``off-policy.'' Q-learning \cite{watkins1992q} is the classic off-policy RL algorithm that estimates the \emph{optimal} action values, which are the action values of an optimal policy, from samples collected by a policy that visits every state-action pair infinitely often. Once we know the approximate optimal action values, a policy greedy with respect to the estimated values would be near-optimal.
If we wish to build continuously adaptive systems that learn and improve online, a more appropriate formulation is regret minimization.

\subsection{Regret minimization}
Suppose that we measure learning progress not by samples but by rewards. We can evaluate how well an algorithm behaves by using the rewards in the environment. In this setting, we compare the rewards collected by a learner as it learns to the rewards would have been collected by a policy optimal for the environment. This is the basis for the game-theoretic notion of regret.
This analysis framework straightforwardly captures the effect of acting on both learning and reward collecting. Good algorithms in such analysis balance exploration and exploitation to learn a good policy efficiently.

\chapter{Efficient policy evaluation}\label{ch:policy-evaluation}

\section{Introduction}

The problem of policy evaluation arises naturally in the context of reinforcement learning (RL) \citep{sutton2018reinforcement} when one wants to evaluate the (action) values of a policy in a Markov decision process (MDP). In particular, policy iteration~\citep{howard1960dynamic} is a classic algorithmic framework for solving MDPs that poses and solves a policy evaluation problem during each iteration. Being motivated by the setting of reinforcement learning, i.e., the underlying MDP parameters are unknown and samples are obtained interactively, we focus on solving the policy evaluation problem given only a \emph{single} sample path.

Following a stationary Markov policy in an MDP, i.e., actions are determined based solely on the current state, gives rise to a \emph{Markov reward process} (MRP) \citep{puterman1994markov}. For the rest of the article, we focus on MRPs and consider the problem of estimating the infinite-horizon \emph{discounted} state values of an unknown MRP.

A straightforward approach to policy evaluation is to estimate the parameters of the MRP and then the value by plugging them into the classic Bellman equation \eqref{eq:immediate-bellman}~\citep{bertsekas1996neuro}. We call this the model-based estimator in the sequel. This approach has recently been proven to be minimax-optimal given a generative model~\citep{Pananjady_Wainwright_2019} and it provides excellent estimates of discounted values in the single sample path setting as well (Section~\ref{sec:numerical}).
However, model-based estimators suffer from a space complexity of $O\rbk*{S^2}$, where $S$ is the number of states in the MRP. In contrast, \emph{model-free} methods enjoy a lower space complexity of $O(S)$ by not explicitly estimating the model parameters~\citep{sutton1988learning}, but tend to exhibit greater estimation error.

A popular class of estimators, $k$-step bootstrapping temporal difference or TD(k)%
\footnote{An important variant is TD($\lambda$), but we do not include it in our experiments since there is not a canonical implementation of the idea of estimating $\lambda$-return~\citep{sutton2018reinforcement}. However, any implementation is expected to exhibit similar behaviors as TD(k) with large $k$ corresponding to large $\lambda$~\citep{kearns2000bias}.} %
estimates a state's value based on the estimated values of other states. Like the model-based estimator, TD(k) is based on the classic Bellman equation~\eqref{eq:immediate-bellman}. The key property of the Bellman equation~\eqref{eq:immediate-bellman} is that the estimate of a state's value is tied to the estimates of other states. This makes it hard to study the convergence of a specific state's value estimate in isolation and motivates the traditional analysis approach of generative model in existing literature.

Traditionally, prior works~\citep{kearns1999finite,even2003learning,Azar_Munos_Kappen_2013,Pananjady_Wainwright_2019} first show efficient estimation of \emph{all} state values under the ``generative model'' assumption that we can generate a sample of next states and rewards starting in each states, and then invoke an argument that such a batch of samples can be obtained over a single sample path when all states are visited for at least once, i.e., over cover times.
In this work, we break with the traditional approach by directly studying the convergence of the value estimate of a \emph{single} state over the sample path. The convergence over all states is obtained as a simple consequence of the union bound.
Our key insight is that it is possible to circumvent the general difficulties of non-independent samples in the single sample path setting by recognizing the embedded regenerative structure of an MRP.
We alleviate the reliance on estimates of other states by studying segments of the sample path that start and end in the same state, i.e., \emph{loops}. This results in a novel and simple algorithm we call the \emph{loop estimator} (Algorithm~\ref{alg:loop}) which is a plug-in estimator based on a novel loop Bellman equation \eqref{eq:loop-bellman}. One important consequence is that the loop estimator can estimate the value of a single state with a space complexity of $O(1)$ which neither $TD(k)$ or the model-based estimator can achieve.

We first review the requisite definitions (Section~\ref{sec:preliminaries}) and then propose the loop estimator (Section~\ref{sec:loop-estimator}).
First, we analyze the algorithm's rate of convergence over visits to a single state (Theorem~\ref{thm:visits}).
Second, we study many steps it takes to visit a state. Using the exponential concentration of first return times (Lemma~\ref{lem:return-times}), we relate visits to their waiting times and establish the rate of convergence over steps (Theorem~\ref{thm:steps}).
Lastly, we obtain the convergence in $\ell_\infty$-norm over all states via the union bound as a consequence (Corollary~\ref{cor:inf-norm}).
Besides theoretical analysis, we also compare the loop estimator to several other estimators numerically on a commonly used example (Section~\ref{sec:numerical}).
Finally, we discuss the model-based vs. model-free status of the loop estimator (Section~\ref{sec:discussions}).

Our main contributions in this work are two-fold:
\begin{itemize}
  \item By recognizing the embedded regenerative structure in MRPs, we derive a new Bellman equation over loops, segments that start and end in the same state.
  \item We introduce \emph{loop estimator}, a novel algorithm that can provably efficiently estimate the discounted values of a single state in an MRP from a single sample path.
\end{itemize}

In the interest of a concise presentation, we defer detailed proofs to Appendix~\ref{sec:proofs-loop} with fully expanded logarithmic factors and constants. An implementation of the proposed loop estimator and the presented experiments is publicly available.\footnote{\url{https://github.com/falcondai/loop-estimator}}

\section{Related works}

Much work that formally studies the convergence of value estimators (particularly the TD estimators) relies on having access to independent trajectories that start in \emph{all} states \citep{dayan1994td, even2003learning, jaakkola1994convergence, kearns2000bias}.
This is called a \emph{generative model} or sometimes a parallel sampling model \cite{kearns1999finite}. Given a convergence over batches of generative samples, we still need some reduction arguments to actually obtain a batch of generative (or parallel) samples over the sample path of a MRP.
\citet{kearns1999finite} consider how a set of independent trajectories can be obtained via mixing, i.e., approximately samples from the stationary distribution. This suggests on \emph{average} it takes $O\rbk*{t_\text{mix} / p^*}$-many steps where $t_\text{mix}$ is the expected steps to get close to the stationary distribution ($\nicefrac{1}{4}$ in total variation distance) and $p^*$ is the smallest nonzero probability in the stationary distribution.

This reduction can be improved by considering the steps the chain takes to visit all states at least once, i.e., \emph{cover times}, which is exactly when we have a batch of generative samples. This is an improved reduction in that we can study its convergence rate with high probability instead of the average behavior.
But the cover time of a Markov chain can be quite large: its concentration can be related to that of the hitting times to \emph{all} states. In contrast, for a single state, our results compares favorable scaling with the maximal expected hitting time of that specific state. When running over all states, we see that TD(0) updates the value of a single state thus incurring time complexity of $O(1)$ per step whereas running $S$-many the loop estimators incurs time complexity of $O(S)$ per step to update.
We provide more detailed discussion of cover times n Section~\ref{sec:cover-time}.

To ensure consistency of estimation is at all possible, we assume that the specific state to estimate is positive recurrent (Assumption~\ref{assum:reachability}), otherwise we cannot hope to (significantly) improve its value estimate after the final visit (see Section~\ref{sec:final-visit-example} for an illustrative example). We think that this assumption is reasonable as recurrence is a key feature of many Markov chains and it connects naturally to the online (interactive) setting where we cannot arbitrarily restart the chain. Moreover, this assumption is no stronger than the assumption used in the cover time reduction which assumes that we can repeatedly visit all states. If a resetting mechanism is available, values of transient states can be estimated from values of the recurrent states. Furthermore, in a finite MRP, there is at least one recurrent state due to the infinite length of a trajectory.

Besides the interest in the RL community to study the policy evaluation problem, operation researchers were also motivated to study estimation in order to leverage simulations as a computational tool. In such settings, the restriction of estimating only from a single sample path is usually not a concern.
Classic work in simulations by \citet{fox1989simulating} deals with estimating discounted value in a continuous time setting, including an estimator using regenerative structure. In comparison to their work, we provides an instance-dependent rate based on the transition structure which is relevant for the single sample path setting. \citet{Haviv_Puterman_1992} and \citet{derman1970finite} propose unbiased value estimators whereas the loop estimator is biased due to inversion.

Outside of the studies on reward processes, the regenerative structure of Markov chains has found application in the \emph{local} computation of PageRank~\citep{lee2013approximating}.
We make use of a lemma (Lemma~\ref{lem:return-times}, whose proof is included in the Appendix~\ref{sec:proof-lem-return-times} for completeness) from this work to establish an upper bound on waiting times (Corollary~\ref{cor:waiting-times}).
Furthermore, we provide an example to support why hitting times do not exponentially concentrate over its expectation in general (see Section~\ref{sec:hitting-time-example}).
Similar in spirit to the concept of locality studied by \citet{lee2013approximating}, our loop estimator enables space-efficient estimation of a single state value with a space complexity of $O(1)$ and an error bound without explicit dependency on the size of the state space. As a consequence, the loop estimator can provably estimate the value of a state with a finite maximal expected hitting time even if the state space is infinite.

Recently, an independent work by \citet{subramanian2019renewal} makes a similar observation of the regenerative structure and studies using estimates similar to the loop estimator in the context of a policy gradient algorithm. It provides promising experimental results that complement our novel theoretical guarantees on the rates of convergence. Taken together, these works show that regenerative structure is a promising direction in RL.

\section{Preliminaries}\label{sec:preliminaries}
\subsection{Markov reward processes and Markov chains}

Consider a finite state space $\mathcal{S} \coloneqq \{1, \cdots, S\}$ whose size is $S = \lvert \mathcal{S}\rvert$, %
a transition probability matrix $\mathbf{P} : \mathcal{S} \times \mathcal{S} \rightarrow [0, 1]$ that specifies the transition probabilities between consecutive states $X_t$ and $X_{t+1}$, %
i.e., (strong) Markov property $\Prob[X_{t+1}=s' \vert X_t = s, \cdots, X_0] = \Prob[X_{t+1}=s' \vert X_t=s] = P_{s s'} $, %
and a reward function $r : \mathcal{S} \rightarrow \mathcal{P}([0, r_\text{max}])$ where $R_t \sim r(X_t)$, then $(X_t, R_t)_{t \geq 0}$ is called a discrete-time finite \emph{Markov reward process} (MRP) \citep{puterman1994markov}.
Note that $(X_t)_{t \geq 0}$ is an embedded Markov chain with transition law $\mathbf{P}$. Furthermore, we denote the mean rewards as $\bar{\mathbf{r}} : s \mapsto \Exp[r(s)]$. As conventions, we denote $\Exp_s[\cdot] \coloneqq \Exp[\cdot \vert X_0 = s]$ and $\Prob_s[\cdot] \coloneqq \Prob[\cdot \vert X_0 = s]$.

The first step when a Markov chain visits a state $s$ is called the \emph{hitting time to $s$}, i.e., $H_s \coloneqq \inf \{ t : X_t = s \}$.
Note that if a chain starts at $s$, then $H_s = 0$. We refer to the first time a chain returns to $s$ as the \emph{first return time to $s$}
\begin{equation}\label{eq:first-return-time}
  H_s^+ \coloneqq \inf \{ t > 0 : X_t = s \}.
\end{equation}
\begin{defn}[Expected recurrence time]\label{def:expected-recurrence-time}
  Given a Markov chain, we define the \emph{expected recurrence time of state $s$} as the expected first return time of $s$ starting in $s$
  \begin{equation}\label{eq:expected-recurrence-time}
    \rho_s \coloneqq \Exp_s \left[ H_s^+ \right].
  \end{equation}
\end{defn}

A state $s$ is \emph{positive recurrent} if its expected recurrence time is finite, i.e., $\rho_s < \infty$.

\begin{defn}[Maximal expected hitting time]\label{def:maximal-expected-hitting-time}
  Given a Markov chain, we define the \emph{maximal expected hitting time of state $s$} as the maximal expected first return time over starting states
  \begin{equation}\label{eq:maximal-expected-hitting-time}
    \tau_s \coloneqq \max_{s' \in \mathcal{S}} \Exp_{s'} [ H_s^+ ].
  \end{equation}
\end{defn}

\subsection{Discounted total rewards}
In RL, we are generally interested in some expected long-term rewards that are collected by following a policy. In the infinite-horizon discounted total reward setting, following a Markov policy on an MDP induces an MRP and the \emph{state value} of state $s$ is
\begin{equation}\label{eq:value}
  v(s) \coloneqq \Exp_s \left[ \sum_{t=0}^\infty \gamma^t R_t \right],
\end{equation}
where $\gamma \in [0, 1)$ is the \emph{discount factor}. Note that since the reward is bounded by $r_\text{max}$, state values are also bounded by $\nicefrac{r_\text{max}}{1 - \gamma}$.
A fundamental result relating values to the MRP parameters $(\mathbf{P}, \bar{\mathbf{r}})$ is the Bellman equation for each state $s \in \mathcal{S}$ \citep{sutton2018reinforcement}
\begin{equation}\label{eq:immediate-bellman}
  v(s) = \bar{r}_s + \gamma \sum_{s' \in \mathcal{S}} P_{s s'} v(s').
\end{equation}

%
\subsection{Problem statement}
Suppose that we have a sample path $(X_t, R_t)_{0 \leq t < T}$ of length $T$ from an MRP whose parameters $(\mathbf{P}, \bar{\mathbf{r}})$ are unknown. Given a state $s$ and discount factor $\gamma$, we want to estimate $v(s)$.

\begin{assum}[State $s$ is reachable]\label{assum:reachability}
  We assume state $s$ is reachable from all states, i.e., $\tau_s < \infty$.
\end{assum}

Otherwise there is some non-negligible probability that state $s$ would not be visited from some starting state. This prevents the convergence in probability (in the form of a PAC-style error bound) that we seek (see Section~\ref{sec:final-visit-example} for a counterexample).
\begin{rem}\label{rem:reachability}
  Assumption~\ref{assum:reachability} can be weakened to the assumption that $s$ is positive recurrent and the MRP starts in the recurrent class containing $s$. All following results can be recovered by restricting $\mathcal{S}$ in the definition of $\tau_s$ to the recurrent class containing $s$. However, for ease of presentation, we will adopt Assumption~\ref{assum:reachability} in the rest of the article without loss of generality.
\end{rem}
Note that Assumption~\ref{assum:reachability} implies the positive recurrence of $s$, i.e., $\rho_s < \infty$, by definition, and that the MRP visits state $s$ for infinitely many times with probability 1.

%

\subsection{Renewal theory and loops}
Stochastic processes in general can exhibit complex dependencies between random variables at different steps, and thus often fall outside of the applicability of approaches that rely on independence assumptions. Renewal theory \citep{ross1996stochastic} focuses on a class of stochastic processes where the process restarts after a renewal event. Such regenerative structure allows us to apply results from the independent and identical distribution (IID) settings.

In particular, we consider the visits to state $s$ as renewal events and define \emph{waiting times} $W_n(s)$ for $n = 1, 2, \cdots$, to be the number of steps before the $n$-th visit
\begin{equation}\label{eq:waiting-times-def}
  W_n(s) \coloneqq \inf \left\{ w : n \leq \sum_{t=0}^w \Ind[ X_t = s ] \right\},
\end{equation}
and the \emph{interarrival times} $I_n(s)$ to be the steps between the $n$-th and $(n+1)$-th visit
\begin{equation}\label{eq:interarrival-times}
  I_n(s) \coloneqq W_{n+1}(s) - W_n(s).
\end{equation}

\begin{rem}\label{rem:times}
  The random times relate to each other in a few intuitive relations. The waiting time of the first visit is the same as the hitting time $W_1(s) = H_s \leq H_s^+$. Waiting times relate to interarrival times $W_{n+1}(s) = W_1(s) + \sum_{i=1}^n I_i(s)$.
\end{rem}
To justify treating visits to $s$ as renewal events, consider the sub-processes starting at $W_1(s)$ and at $W_2(s)$---both MRPs start in state $s$---due to Markov property of MRP, they are statistical replica of each other. Since segments $(X_t, R_t)_{W_n(s) \leq t < W_{n+1}(s)}$ start and end in the same state, we call them \emph{loops}. It follows that loops are independent of each other and obey the same statistical law. Intuitively speaking, an MRP is (probabilistically) specified by its starting state.

\begin{defn}[Loop $\gamma$-discounted rewards]
  Given a Markov reward process and a positive recurrent state $s$, we define the $n$-th \emph{loop $\gamma$-discounted rewards} as the discounted total rewards over the $n$-th loop
  \begin{equation}\label{eq:partial-sum}
    G_n(s) \coloneqq \sum_{u = 0}^{I_n(s) - 1} \gamma^u R_{W_n(s) + u}.
  \end{equation}
\end{defn}

\begin{defn}[Loop $\gamma$-discount]
  Given a Markov reward process and a positive recurrent state $s$, we define the $n$-th \emph{loop $\gamma$-discount} as the total discounting over the $n$-th loop
  \begin{equation}\label{eq:partial-discount}
    \Gamma_n(s) \coloneqq \gamma^{I_n(s)}.
  \end{equation}
\end{defn}

$\left(I_n(s), G_n(s)\right)_{n > 0}$ forms a regenerative process that has nice independence relations. Specifically, $I_n(s) \Perp I_m(s)$, $G_n(s) \Perp G_m(s)$, and $G_n(s) \Perp I_m(s)$ when $n \neq m$. Furthermore, $\rbk*{ I_n(s) }_{n > 0}$ are identically distributed the same as $H_s^+$ when starting in $s$. Similarly, $\rbk*{ G_n(s) }_{n > 0}$ are identically distributed. Note however that $G_n(s) \nPerp I_n(s)$.


\section{Main results}
\subsection{Bellman equations over loops}\label{sec:bellman}
Given the regenerative process $\left(I_n(s), G_n(s)\right)_{n > 0}$, we derive a new Bellman equation over the loops for state value $v(s)$.

\begin{thm}[Loop Bellman equations]\label{thm:loop-bellman}
  Suppose the expected loop $\gamma$-discount is $\alpha(s) \coloneqq \Exp_s[\Gamma_1(s)]$ and the expected loop $\gamma$-discounted rewards is $\beta(s) \coloneqq \Exp_s[G_1(s)]$, we can relate the state value $v(s)$ to itself
  \begin{equation}\label{eq:loop-bellman}
    v(s) = \beta(s) + \alpha(s) \, v(s).
  \end{equation}
\end{thm}

\begin{rem}
  The key difference between the loop Bellman equations \eqref{eq:loop-bellman} and the classic Bellman equations \eqref{eq:immediate-bellman} is the state values involved. Only state value $v(s)$ appears on the right-hand side of \eqref{eq:loop-bellman}.
\end{rem}

\subsection{Loop estimator}\label{sec:loop-estimator}
We plug in the empirical means for the expected loop $\gamma$-discount $\alpha(s)$ and the expected loop $\gamma$-discounted rewards $\beta(s)$ into the loop Bellman equation \eqref{eq:loop-bellman} and define the $n$-th \emph{loop estimator} for state value $v(s)$
\begin{equation}\label{eq:loop-estimator}
  \hat{v}_n(s) \coloneqq \hat{\beta}_n(s) / (1 - \hat{\alpha}_n(s)),
\end{equation}
where
%
%
  $\hat{\alpha}_n(s) \coloneqq \frac{1}{n} \sum_{i=1}^n \gamma^{I_i(s)}$
%
%
and
%
%
  $\hat{\beta}_n(s) \coloneqq \frac{1}{n} \sum_{i=1}^n G_i(s)$.
%
%
Furthermore, we have visited state $s$ for $(N+1)$ times before step $T$ where $N$ is a random variable that counts the number of loops before step $T$
\begin{equation}\label{eq:last-visit-time}
  N \coloneqq \sup \{ n : W_{n + 1}(s) \leq T \},
\end{equation}
and the estimate $\hat{v}_N(s)$ would be the last estimate before step $T$. Hence, with a slight abuse of notations, we define
\begin{equation}
  \hat{v}_T(s) \coloneqq \hat{v}_N(s).
\end{equation}
By using incremental updates to keep track of empirical means, Algorithm~\ref{alg:loop} implements the loop estimator $\hat{v}_T(s)$ with a space complexity of $O(1)$. Running $S$-many copies of loop estimators, one for each state $s \in \mathcal{S}$, takes a space complexity of $O(S)$. Due to the parallel updates, time complexity is $O(S)$ at each step.

\begin{algorithm}[!tb]
    \caption{Loop estimator (for a specific state)}
    \label{alg:loop}
    \begin{algorithmic}[1]
       \STATE {\bfseries Input:} discount factor $\gamma$, state $s$, sample path $(X_t, R_t)_{0 \leq t < T}$ of some length $T$.
       \STATE {\bfseries Return:} an estimate of the discounted value $v(s)$.

       \STATE Initialize the empirical mean of loop discounts $\hat{\alpha} \leftarrow 0$.
       \STATE Initialize the empirical mean of loop discounted rewards $\hat{\beta} \leftarrow 0$.
       \STATE Initialize the loop count $n \leftarrow 0$.
       \FOR {each loop in $(X_t, R_t)_{0 \leq t < T}$}
         \STATE Increment visit count $n \leftarrow n + 1$.
         \STATE Compute the length of the interarrival time $I_n(s) \leftarrow W_{n+1}(s) - W_n(s)$.
         \STATE Compute the partial discounted sum of rewards, $G_n(s) \leftarrow \sum_{u=0}^{I_n(s) - 1} \gamma^u R_{W_n(s) + u}$.
         \STATE Update the empirical means incrementally, $\hat{\alpha} \leftarrow \frac{1}{n} \gamma^{I_n(s)} + \left(1-\frac{1}{n} \right) \hat{\alpha}$, %
         and $\hat{\beta} \leftarrow \frac{1}{n} G_n(s) + \left(1-\frac{1}{n} \right) \hat{\beta}$.

       \ENDFOR
       \RETURN $\hat{\beta} / (1 - \hat{\alpha})$
    \end{algorithmic}
\end{algorithm}

\subsection{Rates of convergence}\label{sec:convergence-rates}
Now we investigate the convergence of the loop estimator, first over visits, i.e., $\hat{v}_n(s) \xrightarrow{p} v(s)$ as $n \rightarrow \infty$, then over steps, i.e., $\hat{v}_T(s) \xrightarrow{p} v(s)$ as $T \rightarrow \infty$.
By applying Hoeffding bound to the definition of loop estimator \eqref{eq:loop-estimator}, we obtain a PAC-style upper bound on the estimation error.

\begin{thm}[Convergence rate over visits]\label{thm:visits}
  Given a sample path from an MRP $(X_t, R_t)_{t \geq 0}$, a discount factor $\gamma \in [0, 1)$, and a positive recurrent state $s$, with probability of at least $1 - \delta$, the loop estimator converges to $v(s)$
  $$ | \hat{v}_n(s) - v(s) | = O\left( \frac{r_\text{max}}{ (1-\gamma)^2 } \sqrt{\frac{1}{n}\, \log \frac{1}{\delta} } \right). $$
\end{thm}

To determine the convergence rate over steps, we need to study the concentration of waiting times which allows us to lower-bound the random visits with high probability. As an intermediate step, we use the fact that the tail of the distribution of first return times is upper-bounded by an exponential distribution per the Markov property of MRP~\citep{lee2013approximating,aldous1999reversible}.

\begin{lem}[Exponential concentration of first return times \citep{lee2013approximating,aldous1999reversible}]\label{lem:return-times}
  Given a Markov chain $(X_t)_{t \geq 0}$ defined on a finite state space $\mathcal{S}$, for any state $s \in \mathcal{S}$ and any $t > 0$, we have
  $$ \Prob \left[ H_s^+ \geq t \right] \leq e \cdot e^{-t / e \tau_s}. $$
\end{lem}

Secondly, since by Remark~\ref{rem:times} we have $W_{n+1}(s) = W_1(s) + \sum_{i=1}^n I_i(s)$, we apply the union bound to upper-bound the tail of waiting times.

\begin{cor}[Upper bound on waiting times]\label{cor:waiting-times}
  With probability of at least $1 - \delta$, $W_n(s) = O\left(n\, \tau_s \, \log \frac{n}{\delta}\right)$.
\end{cor}

Note that the waiting time $W_n(s)$ is nearly linear in $n$ with a dependency on the Markov chain structure via the maximal expected hitting time of $s$, namely $\widetilde{O}(n\, \tau_s)$. In contrast, the \emph{expected} waiting time scales with the expected recurrence time $\Exp [W_n(s)] = \Theta(n\, \rho_s)$. However, an exponential concentration with the expected recurrence time is not possible in general (see Section~\ref{sec:hitting-time-example} for a counterexample).

Using Lambert W function, we invert Corollary~\ref{cor:waiting-times} to lower-bound the visits by step $T$ with high probability. Finally, the convergence rate of $\hat{v}_T(s)$ follows from Theorem~\ref{thm:visits}.

\begin{thm}[Convergence rate over steps]\label{thm:steps}
  With probability of at least $1 - \delta$, for any $T > e \, \delta \, \tau_s$, the MRP $(X_t, R_t)_{0 \leq t < T}$ visits state $s$ for at least $\widetilde{\Omega}(T/\tau_s)$ many times, and the last loop estimate converges to $v(s)$
  $$ \vbk*{ \hat{v}_T(s) - v(s) } = \widetilde{O}\, \rbk*{ \frac{r_\text{max}}{(1-\gamma)^2} \sqrt{\frac{\tau_s}{T} \log \frac{1}{\delta}} } .$$
\end{thm}

Suppose we run a copy of loop estimator to estimate each state's value in $\mathcal{S}$, and denote them with a vector $\hat{\mathbf{v}}_T : s \mapsto \hat{v}_T(s)$. Convergence of the estimation error $\hat{\mathbf{v}}_T - \mathbf{v}$ in terms of the $\ell_\infty$-norm follows immediately by applying the union bound.

\begin{cor}[Convergence rate over all states]\label{cor:inf-norm}
  With probability of at least $1 - \delta$, for any $T > e \, \delta \, \max_s \tau_s$, the MRP $(X_t, R_t)_{0 \leq t < T}$ visits each state $s$ for at least $\widetilde{\Omega}(T/\tau_s)$ many times, and the last loop estimates converge to state values $\mathbf{v}$
  $$ \dvbk{ \hat{\mathbf{v}}_T - \mathbf{v} }_\infty = \widetilde{O}\, \rbk*{ \frac{r_\text{max}}{(1-\gamma)^2} \sqrt{\frac{\max_s \tau_s}{T} \log \frac{S}{\delta}} } .$$
\end{cor}

\section{Additional results}

\subsection{Conditions for consistency}
\label{sec:final-visit-example}
We provide an example to show that if a state is not positive recurrent, i.e., transient, then we cannot attain a consistent estimate of its value in general. This suggests that Assumption~\ref{assum:reachability} is not too strong as a sufficient condition to study. Recall that we are interested in consistent estimation of the discounted value of a state given a single sample path from an unknown MRP. If a state is not positive recurrent, then without assuming any reset mechanisms, it is visited finitely many times over any sample path almost surely.

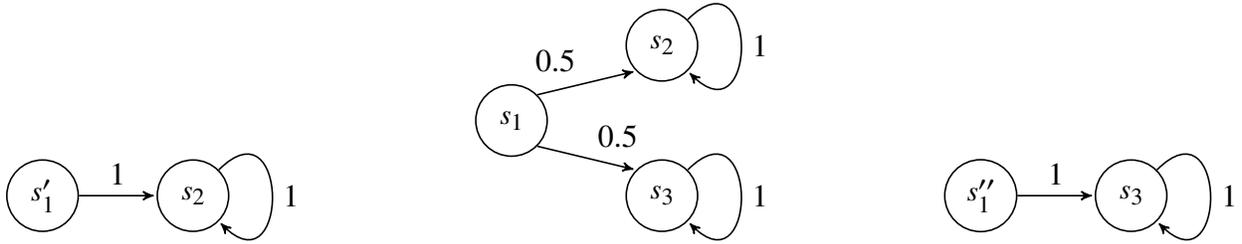
\begin{figure}[!t]
  \centering

    \begin{tikzpicture}[->, >=stealth', shorten >=1pt, auto, semithick]
      \tikzstyle{action} = [draw=black,fill=none]
        \node[state, scale=0.9] at (0, 0) (s1) {$s_1'$};
        \node[state, scale=0.9] at (2, 0) (s2) {$s_2$};

        \path (s1) edge [out=0, in=180] node {$1$} (s2);
        \path (s2) edge [out=45, in=-45, looseness=5] node {$1$} (s2);
    \end{tikzpicture}
  %
  \hfill
  %
    \begin{tikzpicture}[->, >=stealth', shorten >=1pt, auto, semithick]
      \tikzstyle{action} = [draw=black,fill=none]
        \node[state, scale=0.9] at (0, 0) (s1) {$s_1$};
        \node[state, scale=0.9] at (2, 1) (s2) {$s_2$};
        \node[state, scale=0.9] at (2, -1) (s3) {$s_3$};

        \path (s1) edge [out=45, in=-135, looseness=0] node {$0.5$} (s2)
                   edge [out=-45, in=135, looseness=0] node {$0.5$} (s3);
        \path (s2) edge [out=45, in=-45, looseness=5] node {$1$} (s2);
        \path (s3) edge [out=45, in=-45, looseness=5] node {$1$} (s3);
    \end{tikzpicture}
  %
  \hfill
  %
    \begin{tikzpicture}[->, >=stealth', shorten >=1pt, auto, semithick]
      \tikzstyle{action} = [draw=black,fill=none]
        \node[state, scale=0.9] at (0, 0) (s1) {$s_1''$};
        \node[state, scale=0.9] at (2, 0) (s3) {$s_3$};

        \path (s1) 
                   edge [out=0, in=180] node {$1$} (s3);
        \path (s3) edge [out=45, in=-45, looseness=5] node {$1$} (s3);
    \end{tikzpicture}
  \caption{Diagram of three Markov reward processes with transition probabilities labeled on the edges. The rewards are $1$ for $s_2$ and $0$ elsewhere.}
  \label{fig:final-visit-example}
\end{figure}

Consider the following three MRPs in Figure~\ref{fig:final-visit-example}. It is obvious that $v(s_1') = \nicefrac{\gamma}{1 - \gamma}$, $v(s_1'') = 0$, and $v(s_1) = \nicefrac{\gamma}{2(1 - \gamma)}$.
Suppose we start in $s_1$ (of the MRP in the middle), there are only two possible sample paths: $(s_1, 0, s_2, 1, s_2, 1, \cdots)$ and $(s_1, 0, s_3, 0, s_3, 0, \cdots)$.
Note that $s_1$ is only visited \emph{once} in either sample path thus transient.
Furthermore, we obtain the first sample path with probability of $\nicefrac{1}{2}$ in which case we cannot distinguish it from a sample path from the MRP on the top. Similarly, with probability of $\nicefrac{1}{2}$, we get the second sample path which is indistinguishable from a sample path from the MRP at the bottom. However, the values of $s_1'$ and $s_1''$ are different (and both not equal to that of $s_1$). Hence we cannot devise an estimator that can \emph{consistently} estimate $v(s_1)$, $v(s_1')$ and $v(s_1'')$.

\subsection{Concentration of first return times}
\label{sec:hitting-time-example}
We provide an example to show that an exponential concentration of first return times given the expected recurrence time is impossible. In contrast, in Lemma~\ref{lem:return-times} we proved an exponential concentration given the maximal expected hitting time. Furthermore, this is consistent with what one would expect from Markov's inequality since first return times are nonnegative random variables.

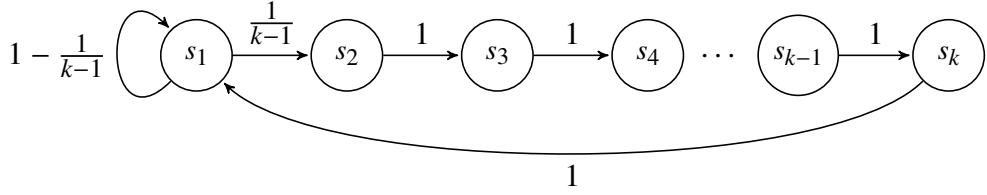
\begin{figure}[!t]
  \centering
  \begin{tikzpicture}[->, >=stealth', shorten >=1pt, auto, semithick]
    \tikzstyle{action} = [draw=black,fill=none]
      \node[state, scale=0.9] at (0, 0) (s1) {$s_1$};
      \node[state, scale=0.9] at (2, 0) (s2) {$s_2$};
      \node[state, scale=0.9] at (4, 0) (s3) {$s_3$};
      \node[state, scale=0.9] at (6, 0) (s4) {$s_4$};
      \node[state, scale=0.9] at (8, 0) (s5) {$s_{k-1}$};
      \node[state, scale=0.9] at (10, 0) (s6) {$s_k$};

      \path (s1) edge [out=0, in=180] node {$\frac{1}{k-1}$} (s2)
                 edge [out=-135,in=135,looseness=5] node {$1 - \frac{1}{k-1}$} (s1);
      \path (s2) edge [out=0, in=180] node {$1$} (s3);
      \path (s3) edge [out=0, in=180] node {$1$} (s4);
      \path (s4) -- node[auto=false]{$\dots$} (s5);
      \path (s5) edge [out=0, in=180] node {$1$} (s6);
      \path (s6) edge [out=-135, in=-45,looseness=0.5] node {$1$} (s1);
  \end{tikzpicture}
  \caption{Diagram of Markov chain $M_k$ with transition probabilities labeled on the edges.}
  \label{fig:hitting-time-example}
\end{figure}

Consider a class of Markov chains $\cbk*{M_k}$ indexed by $k \ge 3$ where Markov chain $M_k$ has a state space $\cbk*{s_1, \cdots, s_k}$ and a transition kernel as depicted in Figure~\ref{fig:hitting-time-example}. Starting in $s_1$, the chain $M_k$ can either transition back to $s_1$ in one step with probability $1 - \frac{1}{k - 1}$ or to $s_2$ with probability $\frac{1}{k-1}$.
Thus, there are only two possible values for the first return time to $s_1$: 1 by the self-transition, and $k$ by going through $s_2, s_3, \ldots, s_k, s_1$.
We can calculate the expected recurrence time as
$$ \rho_{s_1} = \rbk*{1 - \frac{1}{k-1}} \cdot 1 + \frac{1}{k-1} \cdot k = 2 . $$
Suppose that there is an exponential concentration of $H^+_{s_1}$ given $\rho_{s_1}$, then we can upper-bound $\Prob\sbk*{H^+_{s_1} \ge t}$ by some exponential function of $t$. However $H^+_{s_1} = k$ with probability of $\frac{1}{k-1}$ in $M_k$ makes such an exponential bound impossible as the same $\rho$-dependent upper bound has to work for all $M_k$.

Note that the maximal expected hitting time is $\tau_{s_1} = k - 1$ in $M_k$ when the chain starts in $s_2$. A $\tau$-dependent upper bound such as Lemma~\ref{lem:return-times} is not denied by such example.

\subsection{Cover times}
\label{sec:cover-time}

To convert convergence results obtained under the generative model to per steps, we can consider the concentration of cover times.

\begin{defn}[Cover time]
  Given a Markov chain $(X_t)_{t \ge 0}$ with state space $\states$, \emph{cover time} is a stopping time when we have visited all the states at least once
    $$C \coloneqq \inf \cbk*{ t : \{X_0, \dots, X_t\} = \states }.$$
\end{defn}

Conveniently, we can rewrite cover times as the maximum of the hitting times to all the states
$$ C = \max_{s \in \states} H_s. $$

As a consequence, we can apply Lemma~\ref{lem:return-times} and obtain a concentration of cover times using the maximal expected hitting time.
\begin{pro}
  Given a Markov chain $(X_t)_{t \geq 0}$ defined on a finite state space $\mathcal{S}$, with probability of at least $1 - \delta$
  $$C \leq e \max_s \tau_s \log \frac{e S}{\delta}.$$
\end{pro}

Using this upper bound, convergence results obtained under generative model reduction such as for the model-based estimator \cite{Pananjady_Wainwright_2019} match the statistical rate of $\widetilde{O}\rbk*{\sqrt{\nicefrac{\max_s \tau_s}{T}}}$ of the $\ell_\infty$ bound of the loop estimator (Corollary~\ref{cor:inf-norm}). It should be noted that the loop estimator
retains its advantage in local value estimation due to the state-specific constant.

\section{Numerical experiments}\label{sec:numerical}
\begin{figure}[!tb]
  \centering
        \begin{tikzpicture}[->, >=stealth', shorten >=1pt, auto, semithick]
          \tikzstyle{action} = [draw=black,fill=none]


            \node[state, scale=0.9] at (0, 0) (s1) {$s_1$};
            \node[state, scale=0.9] at (1 * 3, 0) (s2) {$s_2$};
            \node[state, scale=0.9] at (2 * 3, 0) (s3) {$s_3$};
            \node[state, scale=0.9] at (3 * 3, 0) (s4) {$s_4$};
            \node[state, scale=0.9] at (4 * 3, 0) (s5) {$s_5$};
            \node[state, scale=0.9] at (5 * 3, 0) (s6) {$s_6$};

            %

            \path (s1) edge [out=45, in=135] node {$0.3$} (s2)
                       edge [out=-45,in=-135,looseness=3] node {$0.7$} (s1);
            \path (s2) edge [out=45, in=135] node {$0.3$} (s3)
                       edge [out=-45,in=-135,looseness=3] node {$0.6$} (s2)
                       edge [out=-180, in=0] node {$0.1$} (s1);
            \path (s3) edge [out=45, in=135] node {$0.3$} (s4)
                       edge [out=-45,in=-135,looseness=3] node {$0.6$} (s3)
                       edge [out=-180, in=0] node {$0.1$} (s2);
            \path (s4) edge [out=45, in=135] node {$0.3$} (s5)
                       edge [out=-45,in=-135,looseness=3] node {$0.6$} (s4)
                       edge [out=-180, in=0] node {$0.1$} (s3);
            \path (s5) edge [out=45, in=135] node {$0.3$} (s6)
                       edge [out=-45,in=-135,looseness=3] node {$0.6$} (s5)
                       edge [out=-180, in=0] node {$0.1$} (s4);
            \path (s6) edge [out=-45,in=-135,looseness=3] node {$0.3$} (s6)
                       edge [out=-180, in=0] node {$0.7$} (s5);
        \end{tikzpicture}%
    \caption{The induced \texttt{RiverSwim} MRP. The arrows are labeled with transition probabilities. The rewards are all zero except for state $s_6$, where $r(s_6) = 1$.}
    \label{fig:river-swim}
\end{figure}
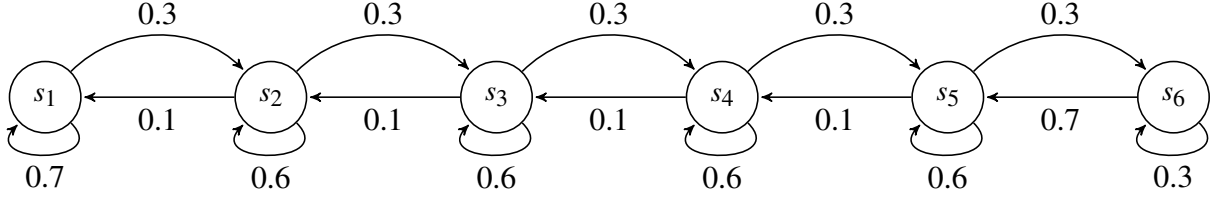

We consider \texttt{RiverSwim}, an MDP proposed by \citet{strehl2008analysis} that is often used to illustrate the challenge of exploration in RL. The MDP consists of six states $\mathcal{S} = \{ s_1, \cdots, s_6 \}$ and two actions $\mathcal{A} = \{\text{``swim downstream''}, \text{``swim upstream''} \}$. Executing the ``swim upstream'' action often fails due to the strong current, while there is a high reward for staying in the most upstream state $s_6$. For our experiments, we use the MRP induced by always taking the ``swim upstream'' action (see Figure~\ref{fig:river-swim} for numerical details).

\begin{figure}
  \includegraphics[width=\columnwidth]{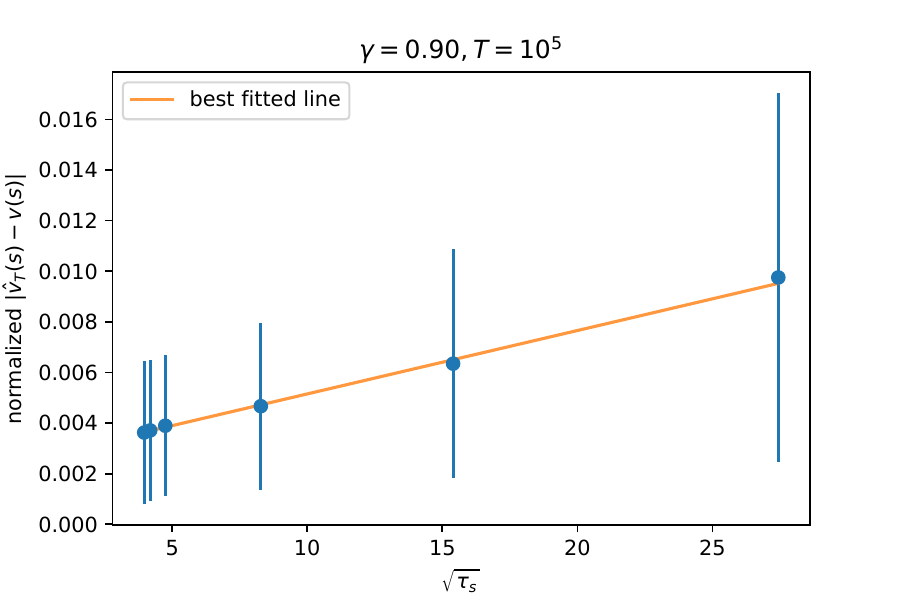}

  \caption{The estimation error of each state (normalized by $\max_s v(s)$) is plotted over the square root of maximal expected hitting times $\sqrt{\tau_s}$ of that state. Error bars show the standard deviations over 200 runs and the discount factor is $\gamma=0.9$ over timesteps $T=10^5$.}
  \label{fig:loop-states}
\end{figure}

The most relevant aspect of the induced MRP is that the maximal expected hitting times are very different for different states: %
$\tau_{s_1} \approx 752$, $\tau_{s_2} \approx 237$, $\tau_{s_3} \approx 68$, $\tau_{s_4} \approx 15$, $\tau_{s_5} \approx 17$, $\tau_{s_6} \approx 22$.
Figure~\ref{fig:loop-states} shows a plot of the estimation errors of the loop estimator for each state over the \emph{square root} of maximal expected hitting times $\sqrt{\tau_s}$ of that state. The observed linear relationship between the two quantities (supported by a good linear fit) is consistent with the instance-dependence in our result of $\vbk*{ \hat{v}_T(s) - v(s) } = \widetilde{O}\rbk*{\sqrt{\tau_s}}$, c.f., Theorem~\ref{thm:steps}.

\subsection{Alternative estimators}
We define several alternative estimators for $v(s)$ and briefly mention their relevance for comparison.

{\bf Model-based.} We compute add-1 smoothed maximum likelihood estimates (MLE) of the MRP parameters $\left(\mathbf{P}, \bar{\mathbf{r}}\right)$ from the sample path
\begin{equation}\label{eq:model-based-p}
  \hat{P}_{s\,s'} \coloneqq \frac{\frac{1}{S} + \sum_{t=0}^{T-1} \Ind \left[ X_{t+1} = s', X_t = s \right]}{1 + \sum_{t=0}^{T-1} \Ind \left[ X_t = s \right]}
\end{equation}
and
\begin{equation}\label{eq:model-based-r}
  \hat{\bar{r}}_s \coloneqq \frac{\sum_{t=0}^{T-1} R_t \Ind \left[ X_t = s \right]}{1 + \sum_{t=0}^{T-1} \Ind \left[ X_t = s \right]}.
\end{equation}
We then solve for the discounted state values from the Bellman equation \eqref{eq:immediate-bellman} for the MRP parameterized by $\left( \hat{\mathbf{P}}, \hat{\bar{\mathbf{r}}} \right)$, i.e., the (column) vector of estimated state values
\begin{equation}\label{eq:model-based}
  \hat{\mathbf{v}}_\text{MB} \coloneqq \left( \mathbf{I} - \gamma \hat{\mathbf{P}} \right)^{-1} \hat{\bar{\mathbf{r}}}
\end{equation}
where $\mathbf{I}$ is the identity matrix.

{\bf TD(k).} $k$-step temporal difference (or $k$-step backup) estimators are commonly recursively defined \citep{kearns2000bias} with TD(0) being a textbook classic \citep{bertsekas1996neuro,sutton2018reinforcement}. Let
$\hat{v}_\text{TD}(0, s) \coloneqq 0$
for all states $s \in \mathcal{S}$. And for $t > 0$
\begin{equation*}\label{eq:td-k}
  \hat{v}_\text{TD}(t, s) \coloneqq \left\{
    \begin{aligned}
      & (1 - \eta_t)\, \hat{v}_\text{TD}(t-1, s) \\
      &\quad + \eta_t \Big( \gamma^0 R_t + \cdots + \gamma^{k} R_{t+k} \\
      &\quad + \gamma^{k+1} \hat{v}_\text{TD}(t-1, X_{t+k+1}) \Big), & \text{if }s = X_t & \\
      & \hat{v}_\text{TD}(t-1, s), & \text{otherwise} &
    \end{aligned}
  \right.
\end{equation*}
where $\eta_t$ is the learning rates. A common choice is to set $\eta_t = 1 / \rbk*{ \sum_{u=0}^{t} \Ind \sbk*{ X_u = s } }$ which satisfies the Robbins-Monro conditions \citep{bertsekas1996neuro}.
But it has been shown to lead to slower convergence than $\eta_t = 1 / \rbk*{ \sum_{u=0}^{t} \Ind \sbk*{ X_u = s } }^d$ where $d \in (\nicefrac{1}{2}, 1)$ \citep{even2003learning}.

It is more accurate to consider TD methods as a large family of estimators each with different choices of $k$, $\eta_t$. Choosing these parameters can create extra work and sometimes confusion for practitioners. Whereas the loop estimator, like the model-based estimator, has no parameters to tune. In any case, it is not our intention to compare with the TD family exhaustively (see more results on TD on \citep{kearns2000bias,even2003learning}). Instead, we compare with $\text{TD}(0)$ and $\text{TD}(10)$, both with $d = 1$, and $\text{TD}(0)^*$ with $d = \nicefrac{1}{2}$.

Convergence of a TD estimator may be complicated by the backing up of errors from statistical estimation. It takes $\widetilde{O}(\nicefrac{1}{1-\gamma})$ updates even in the deterministic setting. In contrast, the model-based estimator and the loop estimator would require no further updates to refine their estimates in the deterministic setting.
In comparison to the generative model-based convergence analysis of TD (theorem 4 and 5 of \citep{even2003learning}), our results on loop estimator is markedly simpler. Furthermore, our results shows explicit (state) instance dependence.

\subsection{Comparative experiments}
\begin{figure}[!htb]
  \centering
    \includegraphics[width=0.8\textwidth]{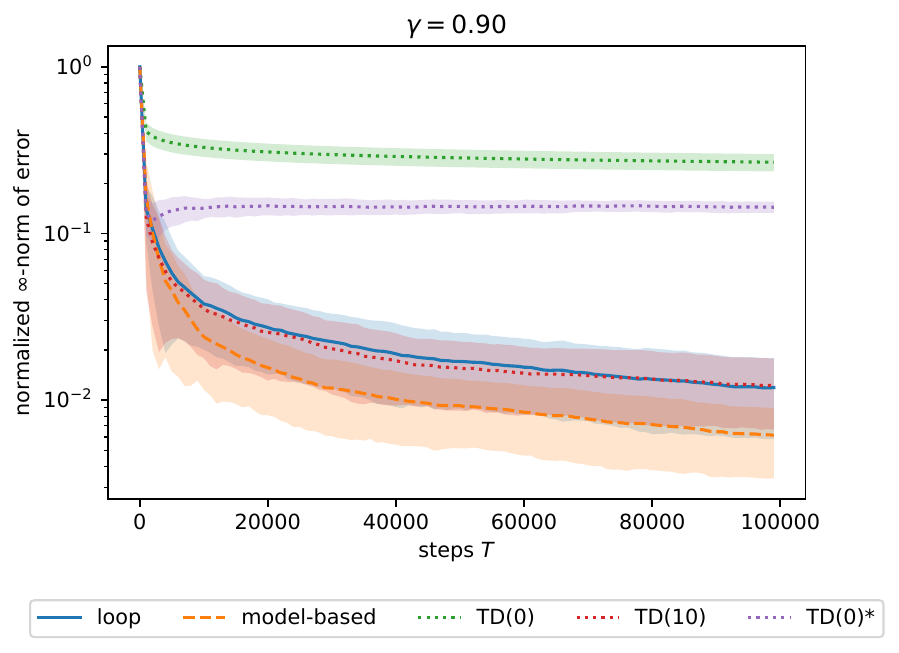}
  \hfill
    \includegraphics[width=0.8\textwidth]{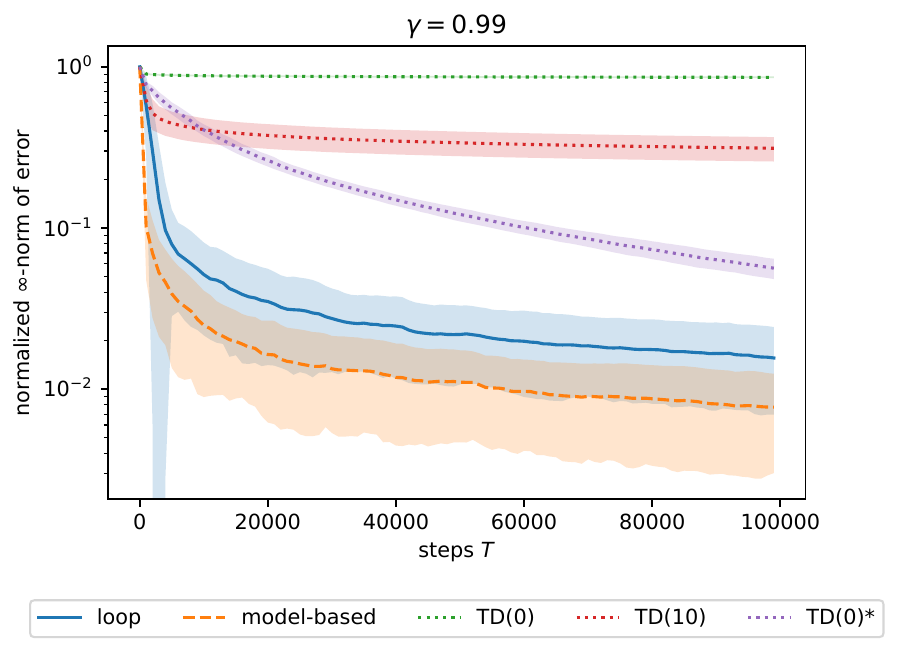}
  \caption{Estimation errors (normalized by $\max_s v(s)$ to be comparable across discount factors) of different estimators at different discount factors (top) $\gamma=0.9$ and (bottom) $\gamma=0.99$. Shaded areas represent the standard deviations over 200 runs. Note the vertical log scale.}
  \label{fig:gammas}
\end{figure}

We experiment with different values for the discount factor $\gamma$, because, roughly speaking, $1 / (1-\gamma)$ sets the horizon beyond which rewards are discounted too heavily to matter. We compare the estimation errors measured in $\ell_\infty$ norm, which is important in RL. The results are shown in Figure~\ref{fig:gammas}.

\begin{itemize}
  \item The model-based estimator dominates all estimators for every discount setting we tested.
  \item TD(k) estimators perform well if $k > 1/(1-\gamma)$. This is related the issue of backing up. TD(k) effectively backs up the value $k$ times at each update and we know that $\widetilde{O}\rbk*{\nicefrac{1}{1 - \gamma}}$-many back ups are needed for TD(0).
  \item The loop estimator performs worse than, but is competitive with, the model-based estimator. Furthermore, similar to the model-based estimator and unlike the TD(k) estimators, its performance seems to be less influenced by discounting.
\end{itemize}
%


\section{Discussions}\label{sec:discussions}
The elementary identity below relates the expected first return times $Y_{s\,s'} \coloneqq \Exp_s \left[ H_{s'}^+ \right]$ to the transition probabilities $P_{s\, s'}$ for a finite Markov chain. Using the matrix notations, suppose that the expected first return times are organized in a matrix $\mathbf{Y}$, and $\mathbf{P}$ the transition matrix of the Markov chain, then we have
$ \mathbf{Y} = \mathbf{P}\left( \mathbf{Y} - \text{diag} \mathbf{Y} + \mathbf{E} \right) $
where $\text{diag} \mathbf{Y}$ is a matrix with the same diagonal as $\mathbf{Y}$ and zero elsewhere, and $\mathbf{E}$ is a matrix with all ones.
Thus, knowing $\mathbf{Y}$ is equivalent to knowing the full model, as we can compute $\mathbf{P}$ using this identity.
Recall that by definition $\Exp\left[ I_1(s) \right] = \Exp_s \left[ H_s^+ \right]$, which is exactly the diagonal of $\mathbf{Y}$.
But only knowing the diagonal is not sufficient to determine the entire set of model parameters, namely $\mathbf{Y}$, the loop estimator based on $\rbk*{I_n}_{n > 0}$ indeed falls short of being a model-based method. It may be considered a \emph{semi}-model-based method as it estimates some but not all of the model parameters.


For large MRPs, a natural extension of our work is to consider recurrence of features instead of states, e.g., a video game screen might not repeat itself completely but the same items might reappear. After all, without repetition exactly or approximately, it would not be possible for an agent to learn and improve its decisions.

We believe that regenerative structure can be further exploited in RL (particularly in the form of the loop Bellman equation \eqref{eq:loop-bellman}) and we think this article provides the fundamental results for future study in this direction.

\chapter{Incorporating expert knowledge}\label{ch:expert-knowledge}

It can be very costly to learn rewarding behavior in an unknown environment. We see that in terms of the samples needed for policy evaluation. In response, we consider ways to incorporate expert knowledge about the environment.
We briefly consider a mixed imitation learning and reinforcement learning \cite{schaff2017towards} where a special query action allows the agent to ask an expert for the optimal action for that state. This is similar to the setting studied by \citet{clouse1996askforhelp}. However, the assumption of knowing the optimal actions for arbitrary states seems too strong rendering the need to reinforcement learn somewhat redundant.
We decide to investigate alternative forms of knowledge that is more compatible to reinforcement learning in the sense that we can still reinforcement learn the optimal behaviors given some ill-specified knowledge. A well-known technique called potential-based reward shaping satisfies such desideratum and more by preserving the linear order over policies by their values (thus suboptimality).

\section{Introduction}
\label{sec:introduction}
In the average reward setting of reinforcement learning (RL) \cite{puterman1994markov,sutton1998reinforcement}, an algorithm learns to maximize its average rewards by interacting with an \emph{unknown} Markov decision process (MDP). Similar to analysis in multi-armed bandits and other online machine learning problems, (cumulative) regret provides a natural model to evaluate the efficiency of a learning algorithm. With the \texttt{UCRL2} algorithm, \citet{jaksch2010near} show a problem-dependent bound of $\tilde{O}(DS\sqrt{AT})$ on regret and an associated logarithmic bound on the expected regret, where $D$ is the diameter of the actual MDP (Definition~\ref{def:diameter}), $S$ the size of the state space, and $A$ the size of the action space. Many subsequent algorithms \cite{fruit2019exploitation} enjoy similar diameter-dependent bounds.
This establishes diameter as an important measure of complexity for an MDP.
However, strikingly, this measure is independent of rewards and is a function of only the transitions.
This is obviously peculiar as two MDPs differing only in their rewards would have the same regret bounds even if one gives the maximum reward for all transitions.
We review the related key observation by \citet{jaksch2010near}, and refine it with a new lemma (Lemma~\ref{lem:kappa-bound}), establishing a reward-\emph{sensitive} complexity measure that we refer to as the \emph{maximum expected hitting cost} (MEHC, Definition~\ref{def:mehc}), which tightens the regret bounds of \texttt{UCRL2} and similar algorithms by replacing diameter (Theorem~\ref{thm:regret-bound}).

Next, with respect to this new complexity measure, we describe a notion of reward informativeness (Section~\ref{sec:informative}). Intuitively speaking, in an environment, the \emph{same} desired policies can be motivated by different (immediate) rewards.
These differing definitions of rewards can be more or less \emph{informative} of useful actions, i.e., yielding high long-term rewards.
To formalize this intuition, we study a way to reparametrize rewards via potential-based reward shaping (PBRS)~\cite{ng1999policy} that can produce different rewards with the same near-optimal policies (Section~\ref{sec:pbrs}). We show that the MEHC changes under reparametrization by PBRS and, in turn, so do regret and sample complexity, substantiating this notion of informativeness. Lastly, we study the extent of its impact. In particular, we show that there is a multiplicative factor-of-two limit on its impact on MEHC in a large class of MDPs (Theorem~\ref{thm:factor-two}). This result and the concept of reward informativeness may be useful for a task designer crafting a reward function (Section~\ref{sec:discussion}). The detailed proofs are deffered to Appendix~\ref{sec:proofs-expert-knowledge}.

The main contributions of this work are two-fold:
\begin{itemize}
  \item We propose a new MDP structural parameter, maximum expected hitting cost (MEHC), that accounts for both transitions and rewards. This parameter replaces diameter in the regret bounds of several model-based RL algorithms.
  \item We show that potential-based reward shaping can change the maximum expected hitting cost of an MDP and thus the regret bound. This results in a set of equivalent MDPs with different learning difficulties as measured by regret. Moreover, we show that their MEHCs differ by a factor of at most two in a large class of MDPs.
\end{itemize}

\subsection{Related work}
\label{sec:related}
This work is closely related to the study of diameter as an MDP complexity measure~\cite{jaksch2010near}, which is prevalent in the regret bounds of RL algorithms in the average reward setting~\cite{fruit2019exploitation}. As noted by \citet{jaksch2010near}, unlike some previous measures of MDP complexity such as the return mixing time~\cite{kearns2002near,brafman2002r}, diameter depends only on the transitions, but not the rewards.
The core reason for the presence of diameter in the regret analysis is that it upper bounds the optimal value span of the extended MDP that summarizes the observations (Section~\ref{sec:ofu} and Equation~\ref{eq:diameter-bound}). We review and update this observation with a reward-dependent parameter we called maximum expected hitting cost (Lemma~\ref{lem:kappa-bound}). Interestingly, the gap between diameter and MEHC can be arbitrarily large $\kappa(M) \leq r_\text{max} D(M)$; there are MDPs with finite MEHC and infinite diameter. These MDPs are non-communicating but have saturated optimal average rewards $\rho^*(M) = r_\text{max}$. Intuitively, there is a state $s$ in these MDPs from which the learner cannot visit some other state $s'$, but can nonetheless achieve the maximum possible average reward, thus allowing for good regret guarantees; the unreachable states do not appear more attractive than the reachable ones under the principle of optimism in the face of uncertainty (OFU).
We will use \texttt{UCRL2} \cite{jaksch2010near} as an example algorithm throughout the rest of the article, however the main results do not depend on it. In particular, with MEHC, its regret bounds are updated (Theorem~\ref{thm:regret-bound}).

Another important comparison is with optimal bias span~\cite{puterman1994markov,bartlett2009regal,fruit2018efficient}, a reward-dependent parameter of MDPs.
Here, we again find that the gap can be arbitrarily large $sp(M) \leq \kappa(M)$.\footnote{This inequality can be derived as a consequence of Lemma~\ref{lem:kappa-bound} as $N(s,a) \rightarrow \infty$, $M^+$ has very tight confidence intervals around the actual transition and mean rewards of $M$. Observe that the span of $u_i$ is equal to $sp(M)$ at the limit of $i \rightarrow \infty$ \cite[remark 8]{jaksch2010near}.} These non-communicating MDPs would have unsaturated optimal average reward $\rho^*(M) < r_\text{max}$.
But as shown elsewhere~\cite{fruit2018near,fruit2018efficient}, extra knowledge of (some upper bound on) the optimal bias span is necessary for an algorithm to enjoy a regret that scales with this smaller parameter. In contrast, \texttt{UCRL2}, which scales with MEHC, does not need to know the diameter or MEHC of the actual MDP.

Potential-based reward shaping \cite{ng1999policy} was originally proposed as a \emph{solution technique} for a programmer to influence the sample complexity of their reinforcement learning algorithm, without changing the near-optimal policies in episodic and discounted settings.
Prior theoretical analysis involving PBRS \cite{ng1999policy,wiewiora2003potential,wiewiora2003principled,asmuth2008potential,grzes2017reward} mostly focuses on the consistency of RL against the shaped rewards, i.e., the resulting learned behavior is also (near-)optimal in the original MDP, while suggesting empirically that the sample complexity can be changed by a well specified potential.
In this work, we use PBRS to construct $\Pi$-equivalent reward functions in the average reward setting (Section~\ref{sec:informative}) and show that two reward functions related by a shaping potential can have different MEHCs, and thus different regrets and sample complexities (Section~\ref{sec:pbrs}).
However, a subtle but important technical requirement of $[0, r_\text{max}]$-boundedness of MDPs makes it difficult to immediately apply our results (Section~\ref{sec:pbrs} and Theorem~\ref{thm:factor-two}) to the treatment of PBRS as a solution technique because an arbitrary potential function picked without knowledge of the original MDP may not preserve the $[0, r_\text{max}]$-boundedness. Nevertheless, we think our work may bring some new perspectives to this topic.

\section{Results}
\label{sec:results}

\subsection{Markov decision process}\label{sec:setting}
A \textit{Markov decision process} is defined by the tuple $M = (\mathcal{S}, \mathcal{A}, p, r)$, where $\mathcal{S}$ is the state space, $\mathcal{A}$ is the action space, $p : \mathcal{S}\times \mathcal{A} \rightarrow \mathcal{P}(\mathcal{S})$ is the transition function, and $r : \mathcal{S} \times \mathcal{A} \rightarrow \mathcal{P}([0, r_\text{max}])$ is the reward function %
with mean $\bar{r}(s, a) \coloneqq \Exp[r(s, a)]$.
We assume that the state and action spaces are finite, with sizes $S \coloneqq \lvert \mathcal{S}\rvert$ and $A \coloneqq \lvert\mathcal{A}\rvert$, respectively.
At each time step $t = 0, 1, 2, \ldots$, an algorithm $\mathfrak{L}$ chooses an action $a_t \in \mathcal{A}$ based on the observations up to that point.
The state transitions to $s_{t+1}$ with probability $p(s_{t+1} \vert s_t, a_t)$ and a reward $r_t \in [0, r_\text{max}]$ is drawn according to the distribution $r(s_t, a_t)$.\footnote{It is important to assume that the support of rewards lies in a \emph{known} bounded interval, often $[0, 1]$ by convention. This is sometimes referred to as a \emph{bounded} MDP in the literature.} Analogous to bandits, the details of the reward distribution often is unimportant, and it suffices to specify an MDP with the mean rewards $\bar{r}$. The transition probabilities and reward function of the MDP are unknown to the learner. The sequence of random variables $(s_t, a_t, r_t)_{t \ge 0}$ forms a stochastic process. Note that a stationary deterministic policy $\pi : \mathcal{S} \rightarrow \mathcal{A}$ is a restrictive type of algorithm whose action $a_t$ depends only on $s_t$. We refer to stationary deterministic policies as policies in the rest of the exposition.

Recall that in a Markov chain, the \emph{hitting time} of state $s'$ starting at state $s$ is a random variable $h_{s \rightarrow s'} \coloneqq \inf \{ t \in \mathbb{N}_{\ge 0} \vert s_t = s' \text{ and } s_0 = s\}$\footnote{$0$-indexing ensures that $h_{s \rightarrow s} = 0$. Note also that by convention, $\inf \varnothing = \infty$.} \cite{levin2008markov}.

\begin{defn}[Diameter, \cite{jaksch2010near}] \label{def:diameter}
Suppose in the stochastic process induced by following a policy $\pi$ in MDP $M$, the time to hit state $s'$ starting at state $s$ is $h_{s \rightarrow s'}(M, \pi)$. We define the \emph{diameter} of $M$ to be
$$ D(M) \coloneqq \max_{s, s' \in \mathcal{S}} \min_{\pi : \mathcal{S} \rightarrow \mathcal{A}} \Exp \left[ h_{s \rightarrow s'}(M, \pi) \right]. $$
\end{defn}

We incorporate rewards into diameter, and introduce a novel MDP parameter.
\begin{defn}[Maximum expected hitting cost] \label{def:mehc}
We define the \emph{maximum expected hitting cost} of a Markov decision process $M$ to be
$$ \kappa(M) \coloneqq \max_{s, s' \in \mathcal{S}} \min_{\pi : \mathcal{S} \rightarrow \mathcal{A}} \Exp \left[ \sum_{t=0}^{h_{s \rightarrow s'}(M, \pi) - 1} r_\text{max} - r_t \right]. $$
\end{defn}
Observe that MEHC is a smaller parameter, that is, $\kappa(M) \leq r_\text{max} D(M)$, since for any $s, s', \pi$, we have $r_\text{max} - r_t \leq r_\text{max}$.

\subsection{Average reward criterion, and regret} \label{sec:regret}
The \textit{accumulated reward} of an algorithm $\mathfrak{L}$ after $T$ time steps in MDP $M$ starting at state $s$ is a random variable
$$ R(M, \mathfrak{L}, s, T) \coloneqq \sum_{t=0}^{T-1} r_t. $$

We define the \textit{average reward} or \textit{gain} \cite{puterman1994markov} as
\begin{equation}\label{eq:criterion}
    \rho(M, \mathfrak{L}, s) \coloneqq \lim_{T \rightarrow \infty} \frac{1}{T} \Exp \left[ R(M, \mathfrak{L}, s, T) \right].
\end{equation}
We evaluate policies by their average reward. This can be maximized by a stationary deterministic policy and we define the \textit{optimal average reward} of $M$ starting at state $s$ as
\begin{equation}\label{eq:opt-avg-reward}
   \rho^*(M, s) \coloneqq \max_{\pi : \mathcal{S} \rightarrow \mathcal{A}} \rho(M, \pi, s).
\end{equation}

Furthermore, we assume that the optimal average reward starting at any state to be the same, i.e., $\rho^*(M, s) = \max_{s'} \rho^*(M, s')$ for any state $s$. This is a natural requirement of an MDP in the online setting to allow for any hope for a vanishing regret. Otherwise, the learner may take actions leading to states with a lower average optimal reward due to ignorance and incur linear regret when compared with the optimal policy starting at the initial state. In particular, this condition is true for communicating MDPs \cite{puterman1994markov} by virtue of their transitions, but this is also possible for non-communicating MDPs with appropriate rewards. We will write $\rho^*(M) \coloneqq \max_{s'} \rho^*(M, s')$.

We compete with the expected cumulative reward of an optimal policy \emph{on its trajectory}, and define the \emph{regret} of a learning algorithm $\mathfrak{L}$ starting at state $s$ after $T$ time steps as
\begin{equation} \label{eq:regret}
   \Delta(M, \mathfrak{L}, s, T) \coloneqq T \rho^*(M) - R(M, \mathfrak{L}, s, T).
\end{equation}

\subsection{Optimism in the face of uncertainty, extended MDP, and \texttt{UCRL2}}
\label{sec:ofu}
The principle of optimism in the face of uncertainty (OFU) \cite{sutton1998reinforcement} states that for uncertain state-action pairs, i.e., those that we have not visited enough up to this point, we should be optimistic about their outcome. The intuition for doing so is that taking reward-maximizing actions with respect to this optimistic model (in terms of both transitions and immediate rewards for these uncertain state-action pairs), we will have no regret if the optimism is well placed or we will quickly learn more about these suboptimal state-action pairs to avoid them in the future. This fruitful idea has been the basis for many model-based RL algorithms \cite{fruit2019exploitation} and in particular, \texttt{UCRL2} \cite{jaksch2010near}, which keeps track of the statistical uncertainty via upper confidence bounds.

Suppose we have visited a particular state-action pair $(s, a)$ $N(s, a)$-many times. With confidence at least $1 - \delta$, we can establish that a confidence interval for both its mean reward $\bar{r}(s, a)$ and its transition $p(\cdot|s, a)$ from the Chernoff-Hoeffding inequality (or Bernstein, \cite{fruit2018near}). Let $b(\delta, n) \in \mathbb{R}$ be the $\delta$-confidence bound after observing $n$ i.i.d. samples of a $[0, 1]$-bounded random variable, $\hat{r}(s, a)$ the empirical mean of $r(s, a)$, $\hat{p}(\cdot|s, a)$ the empirical transition of $p(\cdot|s, a)$. The statistically plausible mean rewards are
$$ B_\delta(s, a) \coloneqq \bigl\{r' \in \mathbb{R} : |r' - \hat{r}(s, a)| \leq r_\text{max} \, b(\delta, N(s, a)) \bigr\} \cap [0, r_\text{max}] $$
and the statistically plausible transitions are
$$ C_\delta(s, a) \coloneqq \bigl\{p' \in \mathcal{P}(\mathcal{S}) : ||p'(\cdot) - \hat{p}(\cdot|s, a)||_1 \leq b(\delta, N(s, a)) \bigr\}. $$

We define an \emph{extended MDP} $M^+ \coloneqq (\mathcal{S}, \mathcal{A}^+, p^+, r^+)$ to summarize these statistics \cite{givan2000bounded,strehl2005theoretical,tewari2007bounded,jaksch2010near}, where $\mathcal{S}$ is the same state space as in $M$, the action space $\mathcal{A}^+$ is a union over state-specific actions
\begin{equation}\label{eq:a-plus}
  \mathcal{A}^+_s \coloneqq \bigl\{ (a, p', r') : a \in \mathcal{A}, p' \in C_\delta(s, a), r' \in B_\delta(s, a) \bigr\},
\end{equation}
where $\mathcal{A}$ is the same action space in $M$, $p^+$ the transitions according to the selected distribution $p'$
\begin{equation}\label{eq:p-plus}
  p^+\big(\cdot|s, (a, p', r') \big) \coloneqq p'(\cdot),
\end{equation}
and $r^+$ is the rewards according to the selected mean reward $r'$
\begin{equation}\label{eq:r-plus}
  r^+\big(s, (a, p', r')\big) \coloneqq r'.
\end{equation}
It is not hard to see that $M^+$ is indeed an MDP with an infinite but compact action space.

By OFU, we want to find an optimal policy for an optimistic MDP within the set of statistically plausible MDPs.
As observed in \cite{jaksch2010near}, this is equivalent to finding an optimal policy $\pi^+ : \mathcal{S} \rightarrow \mathcal{A}^+$ in the extended MDP $M^+$, which specifies a policy in $M$ via $\pi(s) \coloneqq \sigma_1(\pi^+(s))$, where $\sigma_i$ is the projection map onto the $i$-th coordinate (and an optimistic MDP $\widetilde{M} = (\mathcal{S}, \mathcal{A}, \widetilde{p}, \widetilde{r})$ via transitions $\widetilde{p}(\cdot|s,\pi(s)) \coloneqq \sigma_2(\pi^+(s))$ and mean rewards $\widetilde{r}(s, \pi(s)) \coloneqq \sigma_3(\pi^+(s))$ over actions selected by $\pi$\footnote{We can set transitions and mean rewards over actions $a \neq \pi(s)$ to $\hat{p}$ and $\hat{r}$, respectively.}).

By construction of the extended MDP $M^+$, $M$ is in $M^+$ with high confidence, i.e., $\bar{r}(s, a) \in B_\delta(s, a)$ and $p(\cdot|s, a) \in C_\delta(s, a)$ for all $s \in \mathcal{S}, a \in \mathcal{A}$.
At the heart of \texttt{UCRL2}-type regret analysis, there is a key observation \citep[equation (11)]{jaksch2010near} that we can bound the span of optimal values in the \emph{extended} MDP $M^+$ by the diameter of the actual MDP $M$ under the condition that $M$ is in $M^+$.
This observation is needed to characterize how good following the ``optimistic'' policy $\sigma_1(\pi^+)$ in the actual MDP $M$ is.
For $i \geq 0$, the \emph{$i$-step optimal values} $u_i(s)$ of $M^+$ is the expected total reward by following an optimal non-stationary $i$-step policy starting at state $s \in \mathcal{S}$. We can also define them recursively (via dynamic programming\footnote{In fact, the exact maximization of Equation~\ref{eq:u-i} can be found via extended value iteration \citep[section 3.1]{jaksch2010near}})
$$ u_0(s) \coloneqq 0 $$
\begin{align*}
  u_{i+1}(s) &\coloneqq \max_{(a, p', r') \in \mathcal{A}^+_s} \left[ r^+\big(s, (a, p', r')\big) + \sum_{s'} p^+\big(s'|s, (a, p', r')\big) \, u_i(s') \right] \\
  &\triangleright\text{By \eqref{eq:p-plus} and \eqref{eq:r-plus}} \\
  &= \max_{(a, p', r') \in \mathcal{A}^+_s} \left[ r' + \sum_{s'} p'(s') \, u_i(s') \right] \\
  &\triangleright\text{By \eqref{eq:a-plus}} \\
  &= \max_{a \in \mathcal{A}} \left[ \max_{r' \in B_\delta(s, a)} r' + \max_{p' \in C_\delta(s, a)} \sum_{s'} p'(s') \, u_i(s') \right] \label{eq:u-i}\numberthis
\end{align*}

We are now ready to restate the observation. If $M$ is in $M^+$, which happens with high probability, \citet{jaksch2010near} observe that
\begin{equation}\label{eq:diameter-bound}
  \max_s u_i(s) - \min_{s'} u_i(s') \leq r_\text{max} D(M).
\end{equation}

However, this bound is too conservative because it fails to account for the rewards collected. By patching this, we tighten the upper bound with MEHC.

\begin{lem}[MEHC upper bounds the span of values]
\label{lem:kappa-bound}
Assuming that the actual MDP $M$ is in the extended MDP $M^+$, i.e., $\bar{r}(s, a) \in B_\delta(s, a)$ and $p(\cdot|s, a) \in C_\delta(s, a)$ for all $s \in \mathcal{S}, a \in \mathcal{A}$, we have
$$ \max_s u_i(s) - \min_{s'} u_i(s') \leq \kappa(M) $$
where $u_i(s)$ is the $i$-step optimal undiscounted value of state $s$.
\end{lem}

This refined upper bound immediately plugs into the main theorems of \citep[equations 19 and 22, theorem 2]{jaksch2010near}.

\begin{thm}[Reward-sensitive regret bound of \texttt{UCRL2}]
\label{thm:regret-bound}
With probability of at least $1 - \delta$, for any initial state $s$ and any $T > 1$, and $\kappa \coloneqq \kappa(M)$, the regret of \texttt{UCRL2} is bounded by
\begin{align*}
    &\Delta(M, \texttt{UCRL2}, s, T) \\
    &\quad \leq \sqrt{\frac{5}{8} T \log\left(\frac{8T}{\delta}\right)} + \sqrt{T} + \kappa \sqrt{\frac{5}{2} T \log\left(\frac{8T}{\delta}\right)} + \kappa SA \log_2\left( \frac{8T}{SA} \right) \\
    &\quad\quad+ \Bigg(\kappa\sqrt{14 S \log\left(\frac{2AT}{\delta}\right)} + \sqrt{14 \log\left(\frac{2SAT}{\delta}\right)} + 2 \Bigg) (\sqrt{2} + 1) \sqrt{SAT} \\
    &\quad \leq 34 \max\{1, \kappa\} S \sqrt{AT \log\left(\frac{T}{\delta}\right)}.
\end{align*}
\end{thm}

As a corollary, Theorem~\ref{thm:regret-bound} implies that \texttt{UCRL2} offers $O\left( \frac{\kappa^2 S^2 A}{\varepsilon^2} \log \frac{\kappa S A}{\delta \varepsilon} \right)$ sample complexity~\cite{kakade2003sample}, by inverting the regret bound by demanding that the per-step regret is at most $\varepsilon$ with probability of at least $1-\delta$ \citep[corollary 3]{jaksch2010near}.
Similarly, we have an updated logarithmic bound on the expected regret \citep[theorem 4]{jaksch2010near}, $ \Exp [ \Delta(M, \texttt{UCRL2}, s, T) ] = O\rbk*{\frac{\kappa^2 S^2 A \log T}{g}} $ where $g$ is the gap in average reward between the best policy and the second best policy.

\subsection{Informativeness of rewards}
\label{sec:informative}
Informally, it is not hard to appreciate the challenge imposed by delayed feedback inherent in MDPs, as actions with high immediate rewards do not necessarily lead to a high \emph{optimal} value. Are there different but ``equivalent'' reward functions that differ in their \emph{informativeness} with the more informative ones being easier to reinforcement learn? Suppose we have two MDPs differing only in their rewards, $M_1 = (\mathcal{S}, \mathcal{A}, p, r_1)$ and $M_2 = (\mathcal{S}, \mathcal{A}, p, r_2)$, then they have the same diameters $D(M_1) = D(M_2)$ and thus the same diameter-dependent regret bounds from previous works. With MEHC, however, we may get a more meaningful answer.

Firstly, let us make precise a notion of equivalence. We say that $r_1$ and $r_2$ are \emph{$\Pi$-equivalent} if for any policy $\pi : \mathcal{S} \rightarrow \mathcal{A}$, its average rewards are the same under the two reward functions $\rho(M_1, \pi, s) = \rho(M_2, \pi, s)$. Formally, we study the MEHC of a class of $\Pi$-equivalent reward functions related via a potential.

\subsection{Potential-based reward shaping}
\label{sec:pbrs}
Originally introduced by \citet{ng1999policy}, potential-based reward shaping (PBRS) takes a potential $\varphi : \mathcal{S} \rightarrow \mathbb{R}$ and defines shaped rewards
\begin{equation}\label{eq:pbrs}
  r^\varphi_t \coloneqq r_t -\varphi(s_{t}) + \varphi(s_{t+1}).
\end{equation}

We can think of the stochastic process $(s_t, a_t, r^\varphi_t)_{t\geq 0}$ being generated from an MDP $M^\varphi = (\mathcal{S}, \mathcal{A}, p, r^\varphi)$ with reward function $r^\varphi : \mathcal{S} \times \mathcal{A} \rightarrow \mathcal{P}([0, r_\text{max}])$\footnote{One needs to ensure that $\varphi$ respects the $[0, r_\text{max}]$-boundedness of $M$.} whose mean rewards are
$$ \bar{r^\varphi}(s, a) = \bar{r}(s, a) -\varphi(s) + \Exp_{s' \sim p(\cdot|s,a)}\left[ \varphi(s') \right]. $$

It is easy to check that $r^\varphi$ and $r$ are indeed $\Pi$-equivalent. For any policy $\pi$,
\begin{align*}
  \rho(M^\varphi, \pi, s) &= \lim_{T \rightarrow \infty} \frac{1}{T} \Exp \left[ R(M^\varphi, \pi, s, T) \right] \\
    &= \lim_{T \rightarrow \infty} \frac{1}{T} \Exp \left[ \sum_{t=0}^{T-1} r^\varphi_t \right] \\
    &= \lim_{T \rightarrow \infty} \frac{1}{T} \Exp \left[ \sum_{t=0}^{T-1} r_t - \varphi(s_t) + \varphi(s_{t+1}) \right] \\
    &\triangleright\text{By telescoping sums of potential terms over consecutive $t$} \\
    &= \lim_{T \rightarrow \infty} \frac{1}{T} \Exp \left[ - \varphi(s_0) + \varphi(s_T) + \sum_{t=0}^{T-1} r_t \right] \\
    &= \lim_{T \rightarrow \infty} \frac{1}{T} \Big(  - \varphi(s) + \Exp[\varphi(s_T)] + \Exp\left[ R(M, \pi, s, T) \right] \Big) \\
    &\triangleright\text{The first two terms vanish in the limit} \\
    &= \lim_{T \rightarrow \infty} \frac{1}{T} \Exp\left[ R(M, \pi, s, T) \right] \\
    &= \rho(M, \pi, s). \numberthis \label{eq:shaped-avg-reward}
\end{align*}

\begin{figure}[!tb]
  \centering
  \begin{tikzpicture}[->, >=stealth', shorten >=1pt, auto, semithick]
    \tikzstyle{action} = [draw=black,fill=none]
        \node[state] at (-2, 0) (s1) {$s_1$};
        \node[state] at (2, 0) (s2) {$s_2$};

        \node[action, left=of s1] (s1a1) {$a_1$};
        \node[action, above right=of s1] (s1a2) {$a_2$};

        \node[action, right=of s2] (s2a1) {$a_1$};
        \node[action, below left=of s2] (s2a2) {$a_2$};

        \path (s1a2) edge [bend left] node {$\epsilon$} (s2)
                     edge [bend left] node {$1-\epsilon$} (s1);
        \path (s2a2) edge [bend left] node {$\epsilon$} (s1)
                     edge [bend left] node {$1-\epsilon$} (s2);
        \path (s1a1) edge [bend left] node {$1$} (s1);
        \path (s2a1) edge [bend left] node {$1$} (s2);

        \path (s1) edge [-, dashed] node [text=red] {$1-\alpha$} (s1a1);
        \path (s1) edge [-, dashed] node [text=red] {$1-\alpha$} (s1a2);
        \path (s2) edge [-, dashed] node [text=red] {$1-\beta$} (s2a1);
        \path (s2) edge [-, dashed] node [text=red] {$1-\beta$} (s2a2);
    \end{tikzpicture}
  \caption{Circular nodes represent states and square nodes represent actions. The solid edges are labeled by the transition probabilities and the dashed edges are labeled by the mean rewards. Furthermore, $r_\text{max} = 1$. For concreteness, one can consider setting $\alpha = 0.11, \beta = 0.1, \epsilon = 0.05$.}
  \label{fig:example}
\end{figure}
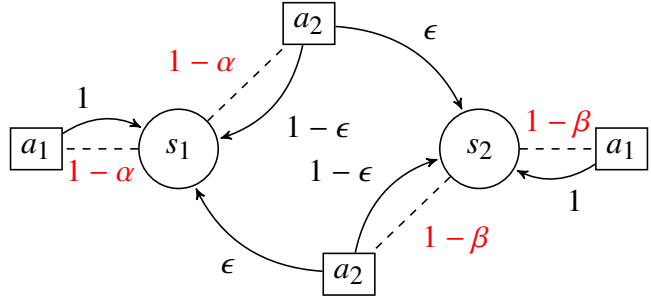

To get some intuition, it is instructive to consider a toy example (Figure~\ref{fig:example}).
Suppose $0 < \beta < \alpha$ and $\epsilon \in (0, 1)$, then the optimal average reward in this MDP is $1 - \beta$, and the optimal stationary deterministic policy is $\pi^*(s_1) \coloneqq a_2$ and $\pi^*(s_2) \coloneqq a_1$, as staying in state $s_2$ yields the highest average reward.
As the expected number of steps needed to transition from state $s_1$ to $s_2$ and vice versa are both $\nicefrac{1}{\epsilon}$ via action $a_2$, we conclude that $\kappa(M) = \max\{ \alpha, \nicefrac{\alpha}{\epsilon}, \nicefrac{\beta}{\epsilon}, \beta \} = \nicefrac{\alpha}{\epsilon}$.
Furthermore, notice that taking action $a_2$ in either state transitions to the other state with probability of $\epsilon$, however the immediate rewards are the same as taking the alternative action $a_1$ to stay in the current state---the immediate rewards are not \emph{informative}.
We can differentiate the actions better by shaping with a potential of $\varphi(s_1) \coloneqq 0$ and $\varphi(s_2) \coloneqq \nicefrac{(\alpha - \beta)}{2 \epsilon}$.
The shaped mean rewards become, at $s_1$,
$$ \bar{r^\varphi}(s_1, a_2) = 1 - \alpha - \varphi(s_1) + \epsilon \varphi(s_2) + (1 - \epsilon) \varphi(s_1) = 1 - \nicefrac{(\alpha + \beta)}{2} > 1 - \alpha = \bar{r^\varphi}(s_1, a_1) $$
and at $s_2$,
$$ \bar{r^\varphi}(s_2, a_2) = 1 - \beta - \varphi(s_2) + \epsilon \varphi(s_1) + (1 - \epsilon) \varphi(s_2) = 1 - \nicefrac{(\alpha + \beta)}{2} < 1 - \beta = \bar{r^\varphi}(s_2, a_1). $$
This encourages taking actions $a_2$ at state $s_1$ and discourages taking actions $a_1$ at state $s_2$ simultaneously. The maximum expected hitting cost becomes smaller
\begin{align*}
    \kappa(M^\varphi) &= \max \left\{ \alpha, \beta, \varphi(s_1) - \varphi(s_2) + \frac{\alpha}{\epsilon}, \, \varphi(s_2) - \varphi(s_1) + \frac{\beta}{\epsilon} \right\} \\
    &= \max \left\{ \alpha, \beta, \frac{\alpha + \beta}{2 \epsilon}, \, \frac{\alpha + \beta}{2 \epsilon} \right\} \\
    &= \frac{\alpha + \beta}{2 \epsilon} \\
    &< \frac{\alpha}{\epsilon} = \kappa(M).
\end{align*}

In this example, MEHC is halved at best when $\beta$ is made arbitrarily close to zero. Noting that the original MDP $M$ is equivalent to $M^\varphi$ shaped with potential $-\varphi$, i.e. $M = (M^\varphi)^{-\varphi}$ from (\ref{eq:pbrs}), we see that MEHC can be almost doubled. It turns out that halving or doubling the MEHC is the most PBRS can do in a large class of MDPs.

\begin{thm}[MEHC under PBRS]
\label{thm:factor-two}
Given an MDP $M$ with finite maximum expected hitting cost $\kappa(M) < \infty$ and an unsaturated optimal average reward $\rho^*(M) < r_\text{max}$, the maximum expected hitting cost of any PBRS-parameterized MDP $M^\varphi$ is bounded by a multiplicative factor of two
$$ \frac{1}{2} \kappa(M) \leq \kappa(M^\varphi) \leq 2 \kappa(M). $$
\end{thm}

The key observation is that the expected total rewards along a loop remains unchanged by shaping, which originally motivated PBRS \cite{ng1999policy}. To see this, consider a loop as a concatenation of two paths, one from $s$ to $s'$ and the other from $s'$ to $s$. Under the shaping of a potential $\varphi$, the expected total rewards of the former is increased by $\varphi(s') - \varphi(s)$ and the latter is decreased by the same amount. For more details, see Appendix~\ref{sec:proof-thm-two}.

\section{Discussion}
\label{sec:discussion}

If we view RL as an engineering tool that ``compiles'' an arbitrary reward function into a behavior (as represented by a policy) in an environment, then a programmer's primary responsibility would be to craft a reward function that faithfully expresses the intended goal. However, this problem of reward design is complicated by practical concerns for the difficulty of learning. As recognized by \citet[section 3.4]{kober2013reinforcement},
\begin{quote}
    ``[t]here is also a trade-off between the complexity of the reward function and the complexity of the learning problem.''
\end{quote}
Accurate rewards are often easy to specify in a sparse manner (reaching a position, capturing the king, etc), thus hard to learn, whereas dense rewards, providing more feedback, are harder to specify accurately, leading to incorrect trained behaviors. The recent rise of deep RL also exposes ``bugs'' in some of these designed rewards \cite{clark2016faulty}. Our results show that the informativeness of rewards, an aspect of ``the complexity of the learning problem'' can be controlled to some extent by a well specified potential without inadvertently changing the intended behaviors of the original reward. Therefore, we propose to separate the definitional concern from the training concern. Rewards should be first defined to faithfully express the intended task, and then any extra knowledge can be incorporated via a shaping potential to reduce the sample complexity of training to obtain the same desired behaviors. That is not to say that it is generally easy to find a helpful potential making the rewards more informative.

Though Theorem~\ref{thm:factor-two} might be a disappointing result for PBRS, we wish to emphasize that this result most directly concerns algorithms whose regrets scale with MEHC, such as \texttt{UCRL2}. It is conceivable that in a different setting such as discounted total rewards, or for a different RL algorithm, such as SARSA with epsilon-greedy exploration \citep[footnote 4]{ng1999policy}, PBRS might have a greater impact on the learning efficiency.

\chapter{Safety, recoverability, and multiple rewards}\label{ch:safety}



We shall first reflect on the idea of acting safely. Sometimes an action can make changes that cannot be undone within the capability of the agent, e.g., dropping and breaking a glass with only some glue at our disposal.
The qualification of capability is important as classical physics is fundamentally reversible albeit sometimes at the cost of a large amount of work.
We can model such situations in a dynamical model such as an MDP by considering whether a state can be reached from another under some policy. In a communicating MDP, by definition, we can reach all the states from all other states (in some steps at some cost). Thus in order to model the absence of full reachability, we need to relax the assumption of communication and consider multichain MDPs.

\section{Irrecoverable MDPs}

More concretely, a multichain MDP has a state space that can be partitioned into components such that we cannot reach some components starting in another component. The actions that take us from one such component to another demand special care since we would not able to return to the first component. So far, the discussion is based on the reachability of states but we would not mind if such an irreversible transition takes place if the second component is as rewarding as the initial one. Some of the irreversible change might in fact be desired, e.g., mixing cream in a cup of coffee. Therefore, we shall focus on the difference in the long-term rewards of the components which is the gain in the average reward setting. We shall refer to a multichain MDP \emph{irrecoverable} if its components do not have the same optimal gain.

\begin{defn}[Recoverability]
  An MDP $\mu$ is \emph{recoverable} if $\max_s g^*(\mu, s) - \min_{s'} g^*(\mu, s') = 0$.
\end{defn}

This concept of safety resembles \emph{controllability} in control theory but a key difference is that we are ultimately concerned with whether rewards can be recovered.
As we have seen previously, a communicating MDP has a constant optimal gain, and thus recoverable.
Another related reward-based concept is robustness where we wish to act defensively against bad luck. This can be formalized by evaluating a policy more conservatively by its cumulative rewards at some percentile rather by the expectation, i.e., its value. Examples are value at risk (VaR) and conditional value at risk (CVaR). Robustness is inherently statistical whereas recoverability is dynamical and thus they are semantically orthogonal to each other.
Value stability is a condition similar to recoverability proposed by \citet{Ryabko_Hutter_2008}. It concerns whether the loss of rewards can be restored in more general processes. They established a general notion of learning and adaptation in such value stable (recoverable) environments. In contrast, we are more interested in notions of learning in an irrecoverable environment.

The regret we used in Chapter~\ref{ch:expert-knowledge} does not work for irrecoverable environments as we cannot hope to achieve sublinear regret due to the presence of states with lower optimal gains. The benchmark used in its definition crucially depends on the MDP being communicating and thus all states have the same optimal gain.
One way to extend the regret over communicating MDPs to multichain MDPs is to account for regret over different sample paths by allowing multiple starts. In such framework, we can think of the multiple starts corresponding to multiple agents acting in replica of the same environment. In such regret over both starts and steps where we would be more concerned with achieving sublinear regret over starts.

Given a set of ongoing trajectories, say $\{ \rbk*{S_t, A_t, R_t}_{t \le t_i} : i = 0, \dots, K \}$, it is not immediately clear which action(s) we get to decide next or if we can decide to start an entirely new trajectory. The different answers to these questions would result in different specifications that should be driven by the applications being modeled. For example, if we think of each trajectory corresponding to a patient in a medical treatment trial, then at each step, we need to decide treatments for all existing patients and a new trajectory is created upon a new patient's arrival. This seems like a demanding scenario. On the opposite end, during robotic controller development in a simulatior, we may be able to decide which simulation to advance and which to pause, and we can decide to spawn another replica altogether.

Even though there can be a variety of learning protocols involving multichain MDPs, we suspect that there is a lower bound on expected regret that is (at least) linear in the number of subchains of the actual MDP. We based this conjecture on lower bounds in MAB by turning each arm into a subchain.


\section{Resets and reset-efficient RL}\label{sec:reset-efficient-rl}


Non-catastrophic mistakes are common and important to deal with in practice as mistakes imply costs. We seek to model such mistakes that cannot be corrected by a robot but can restored by costly interventions of an operator. This is particularly relevant during the development of a controller which aims to avoid catastrophes in deployment.
Developing an optimal controller for an unknown irrecoverable environment is challenging as there are actions that can entrap us in a low rewarding subchain. This sets a harsh limit on learning as it is impossible to behave well in states that we cannot revisit (see Section~\ref{sec:final-visit-example}).

Furthermore, such restrictive learning protocol does not model practical development processes well. During the development of controllers, interventions by human operators (or other equipments) such as replacing excessively worn robot parts or clearing workspaces are routinely performed. The availability of such interventions effectively forgives an agent of mistakes by bringing the system to a specified state by actions beyond the system's capability. Naturally in such setting, we would want to design learning algorithms that utilize these often costly interventions sparingly. We shall call these interventions which restore the system to a distinguished state deterministically \emph{resets}.

Formally, we call an MDP as a \emph{reset-equipped MDP} if it contains a distinguished state $s_i$ called the \emph{initial state} and a distinguished action \texttt{RESET} that deterministically transitions the MDP to $s_i$ with zero reward.

\begin{defn}[reset-equipped MDP]
  Given
  a state space $\mathcal{S}$,
  an \emph{initial state} $s_i \in \states$,
  an action space $\mathcal{A}$,
  a distinguished action $\reset \in \actions$,
  an action sequence $(A_t)_{t \ge 0}$ where $A_t \in \actions$,
  a transition kernel $p : \mathcal{S} \times \mathcal{A} \rightarrow \mathcal{P}\rbk*{\mathcal{S}}$,
  and rewards $r : \mathcal{S} \times \mathcal{A} \rightarrow \mathcal{P}\rbk*{\sbk*{0, \rmax}}$,
  the resulting process $\rbk*{S_t, A_t, R_t}_{t \ge 0}$ is a \emph{reset-equipped MDP} if $\rbk*{S_t, A_t, R_t}_{t \ge 0}$ is an MDP, and that \reset deterministically transition the MDP to the initial state $s_i$ with zero rewards, i.e.,
  $$ \Prob\sbk*{S_{t+1}=s_i \vert S_t=s, A_t=\texttt{RESET}} = 1 \text{ and } \Prob\sbk*{R_t = 0 \vert A_t = \reset} = 1 \text{ for any $s$ and $t$}. $$
\end{defn}

If $\mu = \rbk*{\mathcal{S}, \mathcal{A}, p, r}$ is an irrecoverable MDP and $\tilde\mu = \rbk*{\mathcal{S}, \mathcal{A} \cup \cbk*{\texttt{RESET}}, \tilde{p}, \tilde{r}, s_i}$ an reset-equipped MDP where $\tilde{p}(s, a) = p(s, a)$ and $\tilde{r}(s, a) = r(s, a)$ whenever $a \neq \texttt{RESET}$,
then we say $\tilde\mu$ is reset-augmented from $\mu$, and $\mu$ is reset-restricted from $\tilde\mu$.

The essential challenge of ``irrecoverable'' actions posed to learning is recognized in earlier works \cite{berkenkamp2017safe,moldovan2012safe}.
But prior works \cite{berkenkamp2017safe,moldovan2012safe} define ``recoverability'' more narrowly via transitions, i.e., if a state can be revisited after taking an action, whereas we more broadly consider if the same rewards can still be obtained.
Another limitation is that due to the lack of resets, prior works relied on assumptions of detailed knowledge about the possible environments (through Gaussian process over deterministic dynamics \cite{berkenkamp2017safe}, or Bayesian priors over finite MDPs \cite{moldovan2012safe}) to avoid taking any actions that are risky in any possible environment.
On one hand, this seems overly conservative for an adaptive agent, and on the other hand, it seems to contradict the central tenant of ignorance in learning by demanding detailed prior knowledge about all possible environments.
In summary, without an assumption of some mechanism to ``forgive'' a mistake, these prior studies are more appropriate for modeling the most critical of errors that are truly impossible to recover from in any meaningful sense (even during learning).

By making a mechanism to restart an otherwise irrecoverable MDP available to the algorithm, we can evaluate different learning algorithms by how they utilize resets besides other reward-based metrics such as regret.
The first concern is the choice of benchmark (policy) in this setting. In general, the reset-augmented MDP $\tilde\mu$ has a higher optimal gain than the original MDP $\mu$ due to the additional reset actions. In particular, if $g^*(\tilde\mu, s_i) > g^*(\mu, s_i)$, it implies that all optimal policies have to reset sometimes when starting in $s_i$. This is not the situation we intend to model: the trained policy has no access to resets in deployment. So we shall restrict our attention to suitable MDPs and postpone discussion of the more general case to the next section.

\begin{assum}\label{assum:initial-state-unaffected}
  The optimal gain of the initial state is not increased by the availability of resets
  $g^*(\mu, s_i) = g^*(\tilde\mu, s_i)$.
\end{assum}

We refer to this assumption as \emph{the hazardous setting} and our benchmark policy would be one that achieves the optimal gain on $s_i$ without utilizing any resets. To evaluate learning algorithms, we shall consider both their regret and their use of resets. In fact, we can think of the latter as a regret for the reward function $r(s, a) = -\Ind\sbk*{a=\reset}$.
We have two kinds of rewards and our hazardous setting assumption ensures that the benchmarks, optimal gain and zero reset rate, are achieved by the same policy.
Technically speaking, this assumption keeps us from considering the general setting with multiple reward functions, which we will discuss in Section~\ref{sec:multiple-rewards}.

To avoid technical complications nonessential to the issue at hand such as a disconnected state space, we make some regularity assumptions on the reset-augmented MDPs.

\begin{assum}[One-way reachability]\label{assum:one-way-reachability}
  For any state $s \in \mathcal{S}$, there exists a policy $\pi$ such that we can reach $s$ starting in the initial state $s_i$ with some non-negligible probability.
\end{assum}

\begin{assum}[Non-transient initial state]\label{assum:initial-not-transient}
  The initial state $s_i$ is not a transient state.
\end{assum}

It is not a coincident that we call the special action reset, the same as an API call in the popular OpenAI gym \cite{brockman2016gym}.
In a gym environment and some formulation of MDPs \cite{sutton1998reinforcement}, some states are \emph{absorbing}, meaning that an MDP in those states transitions deterministically to the same states (with zero rewards).
This is a special case of a MDP subchain which contains only a single state. In the gym environment, the environment would inform the agent whether the current state is absorbing via a ``terminal'' indicator variable.
When encountering an absorbing state, the agent is expected to call a ``reset'' API to return to a prescribed initial state or distribution of states. This is exactly what an agent sufficiently certain of itself being in a sub-optimal subchain should do in our proposed framework. Through the lens of comparison, we can think of our model as a generalization of an absorbing state with zero rewards to a subchain with sub-optimal gain which can have a complex MDP structure involving many states and actions. Furthermore, the agent must learn through interactions whether it is in such a sub-optimal subchain.

Specifically, we empirically tested \ucrl, an improved variant of \ucrl with empirical Bernstein bounds \cite{fruit2020improved}, and our proposed algorithm \texttt{Reset-UCRL}.

Since reset-equipped MDPs are communicating, we can directly apply \texttt{UCRL2}. But this is not a good solution as it does not incorporate our knowledge of the \texttt{RESET} actions.

We benefit heavily from \ucrl's explicit representation of the estimated MDP parameters \emph{and} the confidence about them. Specifically, we can supply the true transition probabilities and rewards associated with \reset and eliminate the uncertainty about those parameters in the extended MDP that \ucrl solves in its extended value iteration routine (Algorithm~\ref{alg:reset-ucrl}).

\begin{algorithm}[!tb]
    \caption{Extended value iteration of Reset-UCRL}
    \label{alg:reset-ucrl}
    \begin{algorithmic}[1]
       \STATE {\bfseries Input:} Initial state $s_i$, reset actions, extended MDP, tolerance $\delta$.
       \STATE {\bfseries Return:} Optimistic MDP and its near-optimal policy.

       \STATE Set the empirical transition probabilities of reset actions in the extended MDP to be one if the next state is $s_i$ and zero otherwise.
       \STATE Set the empirical immediate reward of reset actions to be zero.
       \STATE Set the confidence bounds of transitions and rewards of reset actions to be zero.
       \RETURN Run the extended value iteration algorithm with the modified extended MDP and tolerance $\delta$.
    \end{algorithmic}
\end{algorithm}

As we have touched on the connection to gym environments earlier, if the environment provides a terminal signal for some states, we can easily incorporate that information into the estimated model parameters. The explicit representation of the environment parameters (and our confidence) of model-based nature of \ucrl provides intuitive ways to encode our knowledge about the environment.

\begin{figure}
  \includegraphics{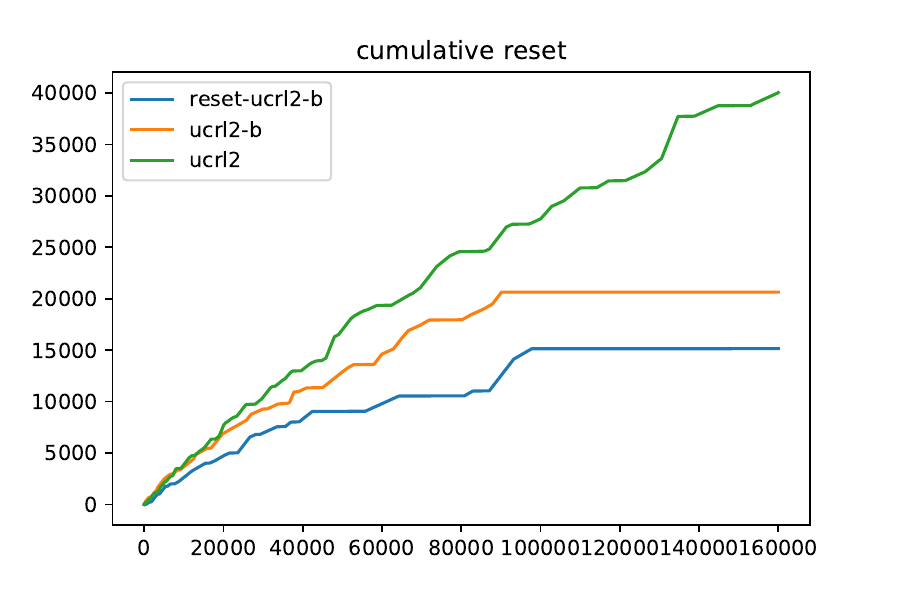}
  \caption{Typical cumulative resets over steps.}
  \label{fig:acc-reset-race}
\end{figure}

\begin{figure}
  \includegraphics{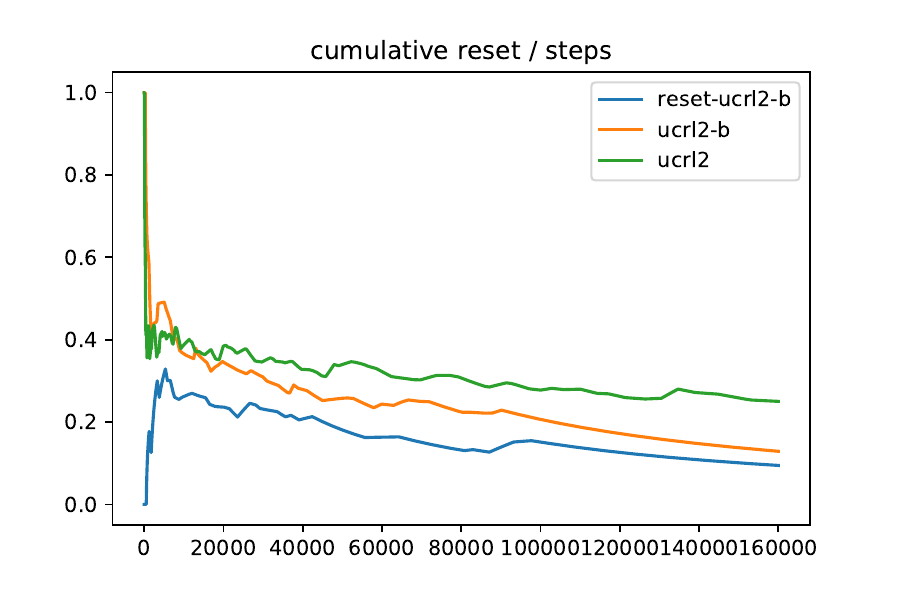}
  \caption{Typical running averages of resets over steps.}
  \label{fig:avg-reset-race}
\end{figure}

\begin{figure}
  \includegraphics{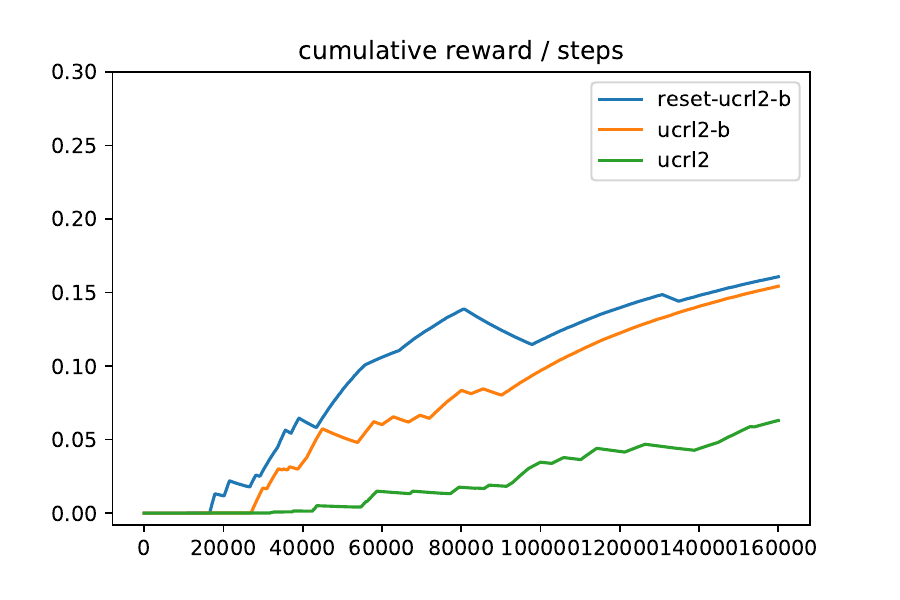}
  \caption{Typical running averages of rewards over steps.}
  \label{fig:avg-reward-race}
\end{figure}

\begin{figure}
  \includegraphics{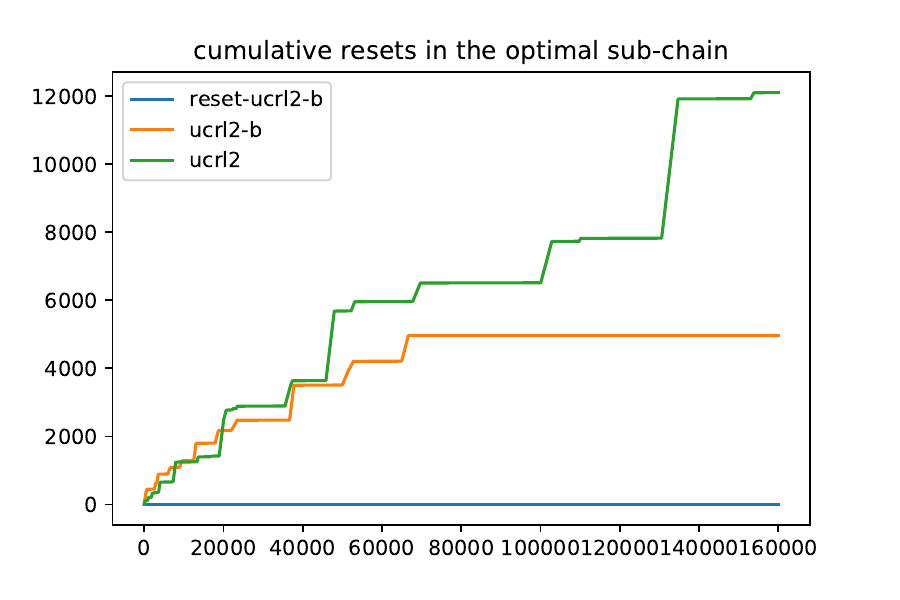}
  \caption{Typical cumulative resets in the optimal sub-chain.}
  \label{fig:acc-opt-chain}
\end{figure}

In numerical experiments\footnote{Details of the MDP we used is deferred to Appendix~\ref{sec:extra-reset-rl}.}, as we expect, all three algorithms steadily improve their average rewards (see Figure~\ref{fig:avg-reward-race}) as their resets plateau (see Figure~\ref{fig:acc-reset-race} and Figure~\ref{fig:avg-reset-race}).
But clearly \texttt{Reset-UCRL} has both superior rewards and resets. (All trajectories share the same random seed.)
More strikingly, we see that \texttt{Reset-UCRL} does not reset at all in the optimal sub-chain (see Figure~\ref{fig:acc-opt-chain}). This is how an agent that knows whether it is in the optimal sub-chain should behave but \texttt{Reset-UCRL} does the same without such knowledge. Its avoidance of unnecessary resetting over these states is a consequence that the uncertainty-free parameters about resets naturally makes them unattractive alternatives.

\section{Planning with multiple rewards}\label{sec:multiple-rewards}

Without the restriction of Assumption~\ref{assum:initial-state-unaffected}, we consider MDPs where resetting can potentially improve the gain of the initial state. This can model problems involving maintenance for example which may tradeoff between productivity and maintenance savings.

A useful observation is that the asymptotic average reset can be written as a negative reward by defining another reward function $r_\text{reset}(s, a) \coloneqq -\Ind\sbk*{a = \texttt{RESET}}$. The negative sign enforces the reward convention and the semantics that fewer resets are preferred. We can define the \emph{reset gain} as the gain with respect to $r_\text{reset}$
$$ \rho(\tilde\mu, \pi, s) \coloneqq \lim_{n\rightarrow\infty} \frac{1}{n} \Exp^\pi_s \sbk*{\sum_{t=1}^n r_\text{reset}\rbk*{S_t, A_t} }. $$

This observation reveals the multi-objective nature of the problem and more specifically, an MDP with two \emph{kinds} of rewards. It is natural to consider an alternative notion of optimality, namely, \emph{(local) Pareto optimality} over the two (or more) objectives, $g(\tilde\mu, \cdot, s_i)$ and $\rho(\tilde\mu, \cdot, s_i)$. Instead of searching for a policy maximizing one of the objectives (or a positive linear combination of them), one might ask for a (locally) Pareto optimal policy.

\begin{defn}[locally Pareto optimality]
  For a metric space $\mathcal{X}$ and a set of (differentiable) objectives $\{ f_i \}$ where $f_i : \mathcal{X} \rightarrow \mathbb{R}$, $x \in \mathcal{X}$ is \emph{locally Pareto-optimal} if there is some neighborhood $N$ around $x$ such that for any $y \in N$, if $f_i(y) > f_i(x)$ for some $i$ then there is some $j$ such that $f_j(y) < f_j(x)$, i.e., $y$ does not dominate $x$.
\end{defn}

In a break from the previous sections, we focus on the planning problem where the MDP parameters $p, r$ are known and we study the search/optimization problem. To setup for the optimization problem, we expand the policy class to be stochastic $\mathcal{S} \rightarrow \mathcal{P}(\mathcal{A})$.

As research interest in RL over increasingly complex environments grows, partly driven by a pursuit of artificial general intelligence, we should recognize that it is unnatural to rank policies linearly in some control problems.
Pareto optimality is natural extension of the optimality of a single objective when there are multiple objectives. It induces a partial order over policies, and the resulting Pareto frontier offers a rich set of ``good'' policies trading off different performance criteria efficiently.
We will continue the discussion of multiple objectives at the more general level of having multiple reward functions mostly ignoring the specifics of the reset gain other than noting that the hazardous setting implies a special Pareto frontier that maximizes two objectives simultaneously.
In prior works, dealing with multiple reward functions is sometimes referred to as a vector-valued MDP \cite{furukawa1980characterization}.


A straightforward metric to quantify the (local) suboptimality of multiple objectives would be $\ell_\infty$ metric which corresponds to the condition that there does not exist any (neighboring) policy that significantly improves some objectives without lowering some. Our goal in planning is to find a near-locally Pareto-optimal policy given the MDP parameters efficiently (in computation).

\begin{figure}
  \includegraphics{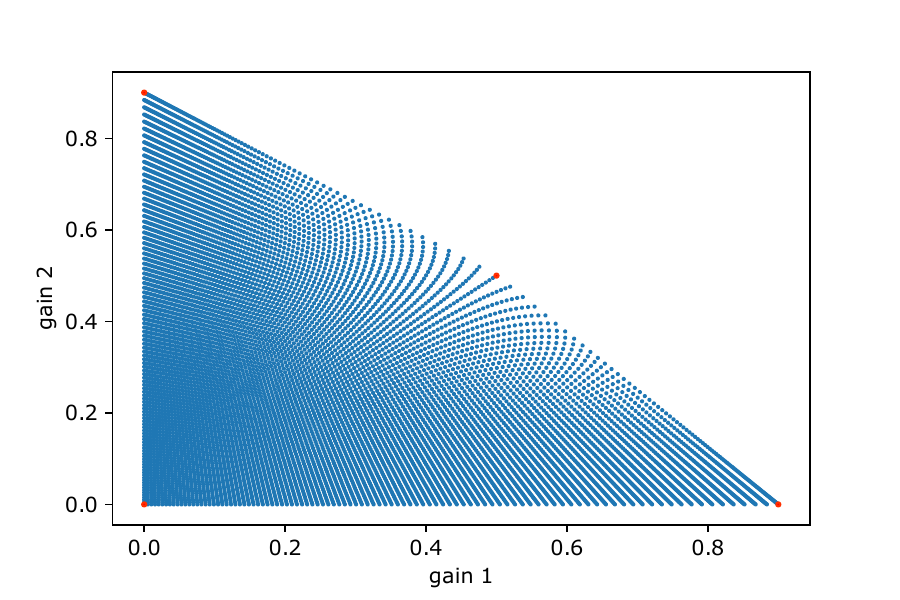}
  \caption{Gains of some stochastic policies (colored in blue) and \emph{all} deterministic policies (colored in red). This illustrates the feasible gains of a particular MDP with two reward functions.}
  \label{fig:feasible-gains}
\end{figure}

In another break from previous problem settings with single-reward MDPs, we need to consider stochastic Markov policies seriously as they can provide some Pareto-efficient solutions with gains that no deterministic Markov policies can attain (see Figure~\ref{fig:feasible-gains} for an example).

The compact solution domain consisted of stochastic policies $\prod_{s \in \states} \mathcal{P}\rbk*{\actions_s}$ naturally suggests local search methods.
There are two primary difficulties in this approach. Firstly, it is not obvious how to efficiently obtain a gradient of the gain with respect to the policy due to the presence of limit in the definition of gain (Definition~\ref{def:gain}). Secondly, after obtaining gradients for each gain, we still need to decide on how to change the policy.

Recall that following a Markov policy in an MDP induces an MRP with transition probabilities and rewards that is weighted by the action probabilities of the policy. The gain can be written as the inner product between the stationary distribution and the rewards. By exploiting a clever observation of \citet{press2012iterated} (in the study of iterated prisoner's dilemma), we can calculate this inner product by computing the determinant of a matrix formed by the transition and the reward of the MRP under some condition (Assumption~\ref{assum:unique-stationary}). The algebraic characterization allows us to differentiate and efficiently compute the exact gradient from a policy. Implementation-wise, this can be done with the help of automatic differentiation software.

\begin{assum}\label{assum:unique-stationary}
  For a stochastic matrix, the multiplicity of its eigenvalue of $1$ is one which implies that it has a unique stationary distribution.
\end{assum}

For an MRP of $S$ states, suppose that its transition matrix is $\mathbf{P} \in \mathbb{R}^{S \times S}$ and the reward is a column vector $\mathbf{r} \in \mathbb{R}^{S \times 1}$.
If $\mathbf{P}$ satisfies Assumption~\ref{assum:unique-stationary}, and its unique stationary distribution is $\mathbf\sigma \in \mathbb{R}^{1 \times S}$, then we can write the inner product as
\begin{equation}\label{eq:inner-prod-det}
  \mathbf\sigma \mathbf{r} = \frac{\det \begin{bmatrix}
    P_{11} - 1 & P_{12}     & \dots  & P_{1(S-1)} & r_{1} \\
    P_{21}     & P_{22} - 1 & \dots  & P_{2(S-1)} & r_{2} \\
    \vdots     & \vdots     & \ddots & \vdots     & \vdots \\
    P_{S1}     & \dots      &        & P_{S(S-1)} & r_{S}
  \end{bmatrix}} {\det \begin{bmatrix}
    P_{11} - 1 & P_{12}     & \dots  & P_{1(S-1)} & 1 \\
    P_{21}     & P_{22} - 1 & \dots  & P_{2(S-1)} & 1 \\
    \vdots     & \vdots     & \ddots & \vdots     & \vdots \\
    P_{S1}     & \dots      &        & P_{S(S-1)} & 1
  \end{bmatrix}}.
\end{equation}

\begin{rem}
  If Assumption~\ref{assum:unique-stationary} does not hold, then we may have different gains depending on which state we start at. Furthermore, the denominator of \eqref{eq:inner-prod-det} would be zero.
\end{rem}


Next, being true to the philosophy of non-linear orderings, we try to find a direction in the policy (tangent) space that improve all gains simultaneously. Such direction exists only if the intersection of all the positive half spaces formed by the gradients is non-empty. If the intersection is empty, we are at a locally Pareto-optimal solution. We find a direction by solving a feasibility linear program by encoding the positive half spaces as linear inequality constraints. The constraints on the search domain (as a subset of $\mathbb{R}^{SA}$) can also be principally incorporated in the linear program as additional linear inequality constraints.

\begin{algorithm}[!tb]
    \caption{Direct-cone policy optimization}
    \label{alg:direct-cone}
    \begin{algorithmic}[1]
       \STATE {\bfseries Input:} multi-reward MDP parameters $\mathbf{P}$, $\{\mathbf{r}_k\}$, and initial stochastic policy $\pi(a|s)$.
       \STATE {\bfseries Return:} a near-locally Pareto-optimal policy.

       \STATE Let the current iterate be the initial policy $\pi$.
       \WHILE {not convergent}
         \STATE Project MDP transitions into an MRP transition matrix with the current policy.
         \FOR {each reward function $\mathbf{r}_k$}
           \STATE Project the MDP reward function into an MRP reward function with the current policy.
           \STATE Compute the gain $g_k$ using \eqref{eq:inner-prod-det}.
           \STATE Differentiate gain $g_k$ with respect to the action probabilities $\pi(a|s)$ to obtain a gradient $\nabla_\pi g_k$.
         \ENDFOR

         \STATE Solve a feasibility linear program to find a vector inside the intersection of positive half spaces induced by gradients $\{\nabla_\pi g_k\}$.
         \IF {not feasible}
           \RETURN the current iterate.
         \ENDIF
         \STATE Line search and increment the current iterate respecting the policy space.
       \ENDWHILE
       \RETURN the current iterate.
    \end{algorithmic}
\end{algorithm}

\begin{figure}
  \includegraphics{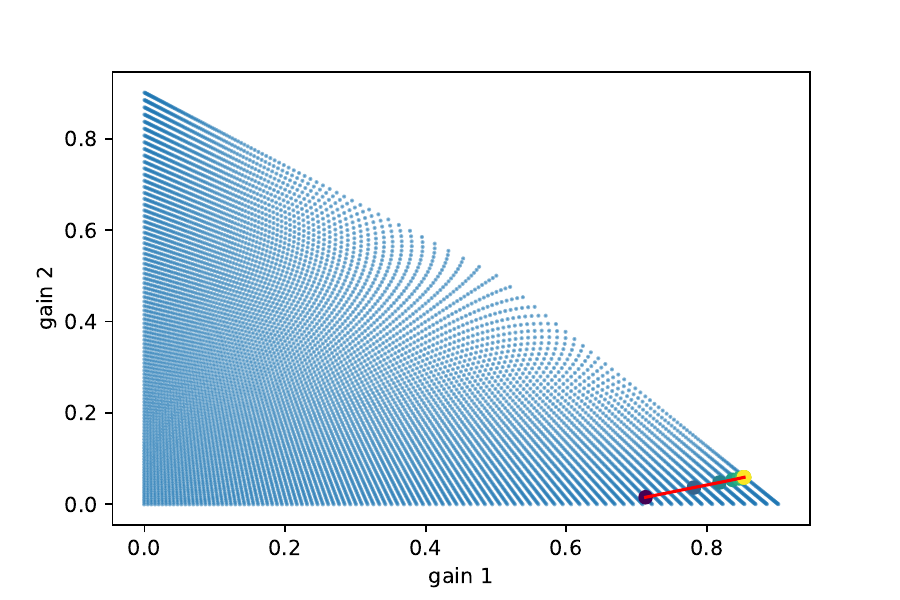}
  \caption{The gains of iterates of a run of the proposed Direct-cone policy optimization algorithm color-coded from dark blue to bright yellow. The blue dots are gains of some stochastic policies as seen in Figure~\ref{fig:feasible-gains}.}
  \label{fig:cramer-gains-t4}
\end{figure}

We can see from a run of the proposed direct-cone policy optimization algorithm, the iterates quickly converged to a policy on the Pareto frontier (see Figure~\ref{fig:cramer-gains-t4}). Notably, the iterates converged to a stochastic policy (compare with Figure~\ref{fig:feasible-gains}). For more details of the numerical experiments, see Appendix~\ref{sec:extra-multiple-reward-exp}.

With the capability of finding a policy on (or near) the Pareto frontier, we can also traverse (close to) the frontier in search of Pareto-efficient policies with different gains. In particular, we can ``steer'' the early iterates towards improving only a subset of gains by excluding the rest from the linear program.


An interesting future work in this direction would be to construct a regret framework that can handle multiple types of regrets. We found it difficult to come up with a sensical regret notion since there are many ``good'' policies in this setting.

\chapter{Concluding thoughts}\label{ch:conclusion}

As we have seen in the results presented, instance-specific bounds often provide insights into what makes an MDP complex. Furthermore, these transition-based and reward-based constants help us understand the behaviors and limitations of our learning algorithms. We think that structures of MDPs deserve more attention in RL. MDPs can be very challenging to learn in general but many MDPs in practice have structures that can be exploited. For example, the loop estimator exploits the fact that we can revisit the same state quickly in some MDPs. It may be possible to efficiently solve MDPs that is simple with respect to some other dynamical constants.

RL algorithms that explicitly represent the MDP parameters may suffer from a high space complexity in comparison to so-called ``model-free'' RL algorithms as in the case of the model-based estimator vs the TD estimators. But as the loop estimator has shown, there may be other ways to be ``model-based'' without estimating all model parameters. By taking the Markov property seriously and estimating a smaller set of parameters, we are able to obtain statistical and computational benefits. In particular, the loop estimator indirectly estimates the transitions by estimating the interarrival times. The connection between transition probabilities and time is a core topic in the study of Markov chains and we believe that such connections may lead to novel RL algorithms.

Besides statistical efficiency, model-based RL algorithms also tend to offers algorithm designers clear ways to encode extra knowledge about the actual MDP if they are available. We exploit this advantage to adapt \ucrl to the setting of reset-equipped MDPs.

While all the works contained herein may benefit from more research, we are particularly excited by the future development of learning protocols of multichain MDPs, of multi-reward MDPs, and RL algorithms specialized to non-generic MDPs. Beyond RL, machine learning problems that involve multiple objectives seem very relevant to practical applications.


\printbibliography

\appendix
\chapter{Appendix}

\allowdisplaybreaks
\section{Detailed proofs of Chapter~\ref{ch:policy-evaluation}}
\label{sec:proofs-loop}

\subsection{Proof of Theorem~\ref{thm:loop-bellman}}
\label{sec:proof-loop-bellman}

\begin{proof}
  Note that since $X_0 = s$, we have $W_1(s) = 0$ and $I_1(s) = W_2(s)$. Since only state $s$ appears here, we will suppress $s$ from the random variables below to simplify the notation. We use Assumption~\ref{assum:reachability} or the weaker assumption that $s$ is positive recurrent, i.e., $\rho_s < \infty$, to guarantee that $W_2 < \infty$ with probability 1.

  \begin{align*}
    &v(s) \\
    &\quad\triangleright\text{By definition \eqref{eq:value}} \\
    &\quad= \Exp_s \left[ \sum_{t=0}^\infty \gamma^t R_t \right] \\
    &\quad\triangleright\text{Split the infinite sum at $W_2$} \\
    &\quad= \Exp_s \left[ \sum_{t=0}^{W_2 - 1} \gamma^t R_t + \sum_{t=W_2}^\infty \gamma^t R_t \right] \\
    &\quad\triangleright\text{Rewrite the indices} \\
    &\quad= \Exp_s \left[ \sum_{t=0}^{W_2 - 1} \gamma^t R_{W_1 + t} + \gamma^{W_2} \sum_{t=0}^\infty \gamma^t R_{W_2 + t} \right] \\
    &\quad\triangleright\text{By definition \eqref{eq:interarrival-times}} \\
    &\quad= \Exp_s \left[ \sum_{t=0}^{I_1 - 1} \gamma^t R_{W_1 + t} + \gamma^{I_1} \sum_{t=0}^\infty \gamma^t R_{W_2 + t} \right] \\
    &\quad\triangleright\text{By definition \eqref{eq:partial-sum}} \\
    &\quad= \Exp_s \left[ G_1 + \gamma^{I_1} \sum_{t=0}^\infty \gamma^t R_{W_2 + t} \right] \\
    &\quad\triangleright\text{Split off the first term} \\
    &\quad= \Exp_s \left[ G_1 \right] + \Exp_s \left[ \gamma^{I_1} \sum_{t=0}^\infty \gamma^t R_{W_2 + t} \right] \\
    &\quad\triangleright\text{Note that $X_{W_2} = s$ and by Markov property} \\
    &\quad= \Exp_s \left[ G_1 \right] + \Exp_s \left[ \gamma^{I_1} \right] \Exp_s \left[ \sum_{t=0}^\infty \gamma^t R_{W_2 + t} \right] \\
    &\quad\triangleright\text{$(R_{W_2+t})_{t \geq 0}$ and $(R_t)_{t \geq 0}$ are probabilistically identical} \\
    &\quad= \Exp_s \left[ G_1 \right] + \Exp_s \left[ \gamma^{I_1} \right] v(s) \\
    &\quad\triangleright\text{By the definitions of $\alpha(s)$ and $\beta(s)$}\\
    &\quad= \beta(s) + \alpha(s) \, v(s). &&\qedhere
  \end{align*}
\end{proof}

\subsection{Proof of Theorem~\ref{thm:visits}}
\label{sec:proof-thm-visits}

\begin{proof}
  Since only state $s$ appears below, we will suppress it in the interest of conciseness. Consider
  \begin{align*}
    &\hat{v}_n - v \\
    &\quad\triangleright\text{By \eqref{eq:loop-bellman} and \eqref{eq:loop-estimator} } \\
    &\quad= \left(\hat{\beta}_n + \hat{\alpha}_n\, \hat{v}_n \right) - \left(\beta + \alpha \, v \right) \\
    &\quad\triangleright\text{Rearrange the terms} \\
    &\quad= \hat{\beta}_n - \beta + \hat{\alpha}_n\, \hat{v}_n - \alpha \, v \\
    &\quad\triangleright\text{Add and subtract $\hat{\alpha}_n\, v$} \\
    &\quad= \hat{\beta}_n - \beta + \hat{\alpha}_n\, \hat{v}_n - \hat{\alpha}_n\, v + \hat{\alpha}_n\, v - \alpha \, v \\
    &\quad= \left(\hat{\beta}_n - \beta \right) + \hat{\alpha}_n \left( \hat{v}_n - v \right) + \left( \hat{\alpha}_n - \alpha \right) v. \\
  \end{align*}

  By the definition of an MRP, we have $v \in [0, \nicefrac{r_\text{max}}{1 - \gamma}]$ and $G_1 \in [0, \nicefrac{r_\text{max}}{1 - \gamma})$. Furthermore, $I_1 \geq 1$ implies that $\gamma^{I_1} \in (0, \gamma]$ and $0 < 1 - \gamma \leq 1 - \hat{\alpha}_n$.
  Hence the estimation error is bounded as follows
  \begin{align*}
    \lvert \hat{v}_n - v \rvert &\leq \lvert \hat{\beta}_n - \beta \rvert + \hat{\alpha}_n \, \lvert \hat{v}_n - v \rvert + \lvert \hat{\alpha}_n - \alpha \rvert \, v \\
    \lvert \hat{v}_n - v \rvert - \hat{\alpha}_n \, \lvert \hat{v}_n - v \rvert &\leq \lvert \hat{\beta}_n - \beta \rvert + \lvert \hat{\alpha}_n - \alpha \rvert \, v \\
    &\triangleright\text{Divide by $1 - \hat{\alpha}_n$} \\
    \lvert \hat{v}_n - v \rvert %
    &\leq \left( 1 - \hat{\alpha}_n \right)^{-1} \, \left( \lvert \hat{\beta}_n - \beta \rvert + \lvert \hat{\alpha}_n - \alpha \rvert \, v \right) \\
    &\leq \frac{1}{1 - \gamma} \, \left( \lvert \hat{\beta}_n - \beta \rvert + \lvert \hat{\alpha}_n - \alpha \rvert \, v \right). \\
  \end{align*}

  With failure probability of at most $\delta/2$, from Hoeffding's inequality, we have
  $$ \lvert \hat{\beta}_n - \beta \rvert < \frac{r_\text{max}}{1 - \gamma} \, \sqrt{ \frac{ \log \nicefrac{4}{\delta} }{2 n} } $$
  and similarly
  $$ \lvert \hat{\alpha}_n - \alpha \rvert < \gamma \, \sqrt{ \frac{ \log \nicefrac{4}{\delta} }{2 n} }. $$
  Applying the union bound, we have
  \begin{align*}
    \lvert \hat{v}_n - v \rvert %
    &\leq \frac{1}{1 - \gamma} \, \left( %
      \frac{r_\text{max}}{1 - \gamma} \, \sqrt{ \frac{ \log \nicefrac{4}{\delta} }{2 n} } %
      + %
      \gamma \, v \, \sqrt{ \frac{ \log \nicefrac{4}{\delta} }{2 n} } %
      \right) \\
    &\leq \frac{1}{1 - \gamma} \, \left( %
      \frac{r_\text{max}}{1 - \gamma} \, \sqrt{ \frac{ \log \nicefrac{4}{\delta} }{2 n} } %
      + %
      \gamma \, \frac{r_\text{max}}{1 - \gamma} \, \sqrt{ \frac{ \log \nicefrac{4}{\delta} }{2 n} } %
      \right) \\
    &< \frac{r_\text{max}}{(1 - \gamma)^2} \, \sqrt{ \frac{ \log \nicefrac{4}{\delta} }{2 n} }. &&\qedhere
  \end{align*}
\end{proof}

\subsection{Proof of Lemma~\ref{lem:return-times}}
\label{sec:proof-lem-return-times}
This proof largely follows the proof by \citet{lee2013approximating} and is presented here in the interest of self-containedness.

\begin{proof}
  Suppose $a, b > 0$, consider the probability of the event that $s$ is not visited in the next $a$ steps given that it is not visited in the previous $b$ steps, that is
  \begin{align*}
    &\Prob \left[ H_s^+ > a + b \middle| H_s^+ > b \right] \\
    &\quad\triangleright\text{In particular, $X_b \neq s$} \\
    &\quad \leq \Prob \left[ H_s^+ > a + b \middle| X_b \neq s \right] \\
    &\quad\triangleright\text{By Markov property, we can shift the index} \\
    &\quad = \Prob \left[ H_s^+ > a \middle| X_0 \neq s \right] \\
    &\quad \leq \max_{s' \in \mathcal{S}} \Prob \left[ H_s^+ > a \middle| X_0 = s' \right] \\
    &\quad\triangleright\text{By Markov inequality} \\
    &\quad \leq \max_{s' \in \mathcal{S}} \frac{\Exp \left[ H_s^+ \middle| X_0 = s' \right]}{a} \\
    &\quad\triangleright\text{By definition \eqref{eq:maximal-expected-hitting-time}} \\
    &\quad \leq \frac{\tau_s}{a}. \label{eq:tau-a} \numberthis
  \end{align*}

  Let $a = e\,\tau_s$, and apply the above $\floor*{\frac{t}{a}}$-many times to
  \begin{align*}
    \Prob \sbk*{H_s^+ \geq t} %
    & \leq \Prob \sbk*{ H_s^+ \geq \floor*{\frac{t}{a}}\, a } \\
    &\triangleright\text{Apply \eqref{eq:tau-a}} \\
    & \leq \rbk*{ \frac{\tau_s}{a} }^{\floor*{ \frac{t}{a} }} \\
    &\triangleright\text{Let $a = e\, \tau_s$} \\
    & \leq \rbk*{ \frac{1}{e} }^{\floor*{\frac{t}{e \tau_s}}} \\
    & < e \cdot \rbk*{\frac{1}{e}}^{\frac{t}{e \tau_s}}. &&\qedhere
  \end{align*}
\end{proof}

\subsection{Proof of Corollary~\ref{cor:waiting-times}}
\label{sec:proof-cor-waiting-times}

\begin{proof}
  For conciseness, we suppress $s$ here since only state $s$ appears.
  Suppose $a > 0$. By Remark~\ref{rem:times}, we have $W_n < n\,a$ if $W_1 < a$ and $I_i < a$ for $i = 1, \cdots, n-1$.
  Note that $W_1 \leq H_s^+$ and $I_i$ distribute identically to $H_s^+$. Immediately from inverting Lemma~\ref{lem:return-times}, we have with failure probability of at most $\nicefrac{\delta}{n}$, $W_1$ is bounded
  $$ W_1 \leq H_s^+ < e \tau_s \log \frac{e n}{\delta}. $$
  Suppose each $I_i < a$ fails with probability of at most $\nicefrac{\delta}{n}$, then we similarly have
  $$ I_i < e \tau_s \log \frac{e n}{\delta}. $$
  Applying the union bound, and with probability of at least $1 - \delta$, we have
  $$ W_n < e n \tau_s \log \frac{e n}{\delta}. $$
\end{proof}

\subsection{Proof of Theorem~\ref{thm:steps}}
\label{sec:proof-thm-steps}
\begin{proof}
  First, we introduce the Lambert $\mathcal{W}$ function to invert Corollary~\ref{cor:waiting-times}.
  Recall that the Lambert $\mathcal{W}$ function is a transcendental function defined such that $\mathcal{W}(x) e^{\mathcal{W}(x)} = x$, and thus it is a monotonically increasing function.
  At step $T$, suppose
  \begin{align*}
    e n\, \tau_s \log \frac{e n}{\delta} &= T \\
    \frac{e n}{\delta}\, \log \frac{e n}{\delta} &= \frac{T}{\delta\,\tau_s} \\
    \rbk*{\log \frac{e n}{\delta}}\, e^{\rbk*{\log \frac{e n}{\delta}}} &= \frac{T}{\delta\,\tau_s} \\
    &\triangleright\text{By the definition of $\mathcal{W}$} \\
    \log \frac{e n}{\delta} &= \mathcal{W}\rbk*{\frac{T}{\delta\,\tau_s}} \\
    &\triangleright\text{Exponentiate both sides} \\
    \frac{e n}{\delta} &= e^{\mathcal{W}\rbk*{\frac{T}{\delta\,\tau_s}}}.
  \end{align*}

  Use the fact that if $x > e$, then $\log x - \log\log x < \mathcal{W}(x)$. So given $T > e \delta \tau_s$, we can lower-bound the number of visits
  \begin{align*}
    e^{ \log \rbk*{\frac{T}{\delta\,\tau_s}} - \log\log \rbk*{\frac{T}{\delta\,\tau_s}} } &< \frac{e n}{\delta} \\
    \frac{ \frac{T}{\delta\,\tau_s} }{\log \rbk*{\frac{T}{\delta\,\tau_s}}} &< \frac{e n}{\delta} \\
    \frac{ T\, }{e\, \tau_s\,\log \rbk*{\frac{T}{\delta\,\tau_s}}} &< n.
  \end{align*}

  Plugging this into Theorem~\ref{thm:visits}, we obtain the desired expression
  $$ \vbk*{\hat{v}_T(s) - v(s)} %
  < \frac{r_\text{max}}{(1 - \gamma)^2} \, \sqrt{ \frac{ e \tau_s\,\log \rbk*{\frac{T}{\delta\,\tau_s}} \log \frac{4}{\delta} }{2 T} }. $$
\end{proof}

\subsection{Proof of Corollary~\ref{cor:inf-norm}}
\label{sec:proof-cor-inf-norm}
\begin{proof}
  We run $S$ many copies of loop estimators, one for each state $s \in \mathcal{S}$.
  Following Theorem~\ref{thm:steps}, with failure probability of at most $\nicefrac{\delta}{S}$, we can ensure that each estimator has an error of at most
  $$ \vbk*{\hat{v}_T(s) - v(s)} %
  < \frac{r_\text{max}}{(1 - \gamma)^2} \, \sqrt{ \frac{ e \tau_s\,\log \rbk*{\frac{S\,T}{\delta\,\tau_s}} \log \frac{4 S}{\delta} }{2 T} }. $$

  The largest upper bound comes from the state with the largest maximal expected hitting time $\max_{s \in \mathcal{S}} \tau_s$ of the Markov chain. Apply the union bound and we have
  $$ \dvbk*{\hat{\mathbf{v}}_T - \mathbf{v}}_\infty %
  < \frac{r_\text{max}}{(1 - \gamma)^2} \, \sqrt{ \frac{e \max_s\tau_s\,\log \rbk*{\frac{S\,T}{\delta\,\min_s\tau_s}} \log \frac{4 S}{\delta} }{2 T} }. $$

\end{proof}

\section{Detailed proofs of Chapter~\ref{ch:expert-knowledge}}\label{sec:proofs-expert-knowledge}
\subsection{Proof of Lemma~\ref{lem:kappa-bound}}
\label{sec:proof-lem-1}

Assuming that the actual MDP $M$ is in the extended MDP $M^+$, i.e., $\bar{r}(s, a) \in B_\delta(s, a)$ and $p(\cdot|s, a) \in C_\delta(s, a)$ for all $s \in \mathcal{S}, a \in \mathcal{A}$, we have
$$ \max_s u_i(s) - \min_{s'} u_i(s') \leq \kappa(M) $$
where $u_i(s)$ is the $i$-step optimal undiscounted value of state $s$.

\begin{proof}
  By assumption, the actual mean rewards $\bar{r}$ and transitions $p$ are contained in the extended MDP $M^+$, i.e., for any $s \in \mathcal{S}$ and $a \in \mathcal{A}$, $\bar{r}(s, a) \in B_\delta(s, a)$ and $p(\cdot|s, a) \in C_\delta(s, a)$.
  Thus for any policy $\pi : \mathcal{S} \rightarrow \mathcal{A}$ in the actual MDP $M$, we can construct a corresponding policy $\pi^+ : \mathcal{S} \rightarrow \mathcal{A}^+$ in the extended MDP $M^+$
  $$ \pi^+(s) \coloneqq \Big( \pi(s), p(\cdot| s, \pi(s)), \bar{r}(s, \pi(s)) \Big). $$
  Following $\pi^+$ in $M^+$ induces the same stochastic process $(s_t, a_t, r_t)_{t \geq 0}$ as following $\pi$ in $M$. In particular they have the same expected hitting times and expected rewards.
  By definition $u_i(s)$ is the value of following an \emph{optimal} $i$-step non-stationary policy starting at $s$ in the extended MDP $\mathcal{M}^+$. For any $s'$, by optimality, $u_i(s)$ must be no worse than first following $\pi^+$ from $s$ to $s'$ and then following the optimal $i$-step non-stationary policy from $s'$ onward. Along the path from $s$ to $s'$, we receive rewards according to $\sigma_3(\pi^+) = \bar{r}$ and after arriving at $s'$, we have missed at most $r_\text{max} h_{s \rightarrow s'}(M^+, \pi^+)$-many rewards of $u_i(s')$ so in expectation
  \begin{align*}
    u_i(s) &\geq \Exp\left[ \sum_{t=0}^{h_{s \rightarrow s'}(M^+, \pi^+) - 1} r_t \right] + u_i(s') - \Exp[r_\text{max} h_{s \rightarrow s'}(M^+, \pi^+)] \\
    &= \Exp\left[ \sum_{t=0}^{h_{s \rightarrow s'}(M^+, \pi^+) - 1} r_t - r_\text{max} \right] + u_i(s') \\
    &\triangleright\text{By definition of $\pi^+$, hitting time $h_{s \rightarrow s'}(M, \pi) = h_{s \rightarrow s'}(M^+, \pi^+)$ } \\
    &= \Exp\left[ \sum_{t=0}^{h_{s \rightarrow s'}(M, \pi) - 1} r_t - r_\text{max} \right] + u_i(s').
  \end{align*}
  Moving the terms around and we get
  $$ u_i(s') - u_i(s) \leq \Exp\left[ \sum_{t=0}^{h_{s \rightarrow s'}(M, \pi) - 1} r_\text{max} - r_t \right]. $$

  Since this holds for any $\pi$ by optimality, we can choose one with the smallest expected hitting cost
  $$ u_i(s') - u_i(s) \leq \min_{\pi : \mathcal{S} \rightarrow \mathcal{A}} \Exp\left[ \sum_{t=0}^{h_{s \rightarrow s'}(M, \pi) - 1} r_\text{max} - r_t \right]. $$

  Since $s, s'$ are arbitrary, we can maximize over pairs of states on both sides and get
  $$ \max_{s'} u_i(s') - \min_s u_i(s) \leq \max_{s, s'} \min_{\pi : \mathcal{S} \rightarrow \mathcal{A}} \Exp\left[ \sum_{t=0}^{h_{s \rightarrow s'}(M, \pi) - 1} r_\text{max} - r_t \right] = \kappa(M). $$

  It should be noted that even in some cases where the hitting time is infinity---in a non-communicating MDPs for example---$\kappa$ can still be finite and this inequality is still true! In these cases, $r_t = r_\text{max}$ except for finitely many terms implying $\rho^*(M, s) = r_\text{max}$.
\end{proof}

\subsection{Proof of Theorem~\ref{thm:factor-two}}
\label{sec:proof-thm-two}

Given an MDP $M$ with finite maximum expected hitting cost $\kappa(M) < \infty$ and an unsaturated optimal average reward $\rho^*(M) < r_\text{max}$, the maximum expected hitting cost of any PBRS-parametrized MDP $M^\varphi$ is bounded by a multiplicative factor of two
$$ \frac{1}{2} \kappa(M) \leq \kappa(M^\varphi) \leq 2 \kappa(M). $$

\begin{proof}
  We denote the expected hitting cost between two states $s, s'$ as
  $$ c(s, s') \coloneqq \min_{\pi : \mathcal{S} \rightarrow \mathcal{A}} \Exp \left[ \sum_{t=0}^{h_{s \rightarrow s'}(M, \pi) - 1} r_\text{max} - r_t \right]. $$

  Suppose that the pair of states $(s, s')$ maximizes the expected hitting cost in $M$ which is assumed to be finite
  $$ \kappa(M) = c(s, s') < \infty. $$
  Furthermore, the condition that $\rho^*(M) < r_\text{max}$ implies that the hitting times are finite for the minimizing policies. This ensures that the destination state is actually hit in the stochastic process.

  Considering the expected hitting cost of the reverse pair, $(s', s)$,
  \begin{equation}\label{eq:max-pair}
    \kappa(M) = \max \{ c(s, s'), c(s', s) \} \leq c(s, s') + c(s', s)
  \end{equation}
  since hitting costs are nonnegative.

  With $\varphi$-shaping,
  \begin{align*}
    c^\varphi(s, s') &= \min_{\pi : \mathcal{S} \rightarrow \mathcal{A}} \Exp \left[ \sum_{t=0}^{h_{s \rightarrow s'}(M, \pi) - 1} r_\text{max} - r^\varphi_t \right] \\
    &= \min_{\pi : \mathcal{S} \rightarrow \mathcal{A}} \Exp \left[ \sum_{t=0}^{h_{s \rightarrow s'}(M, \pi) - 1} r_\text{max} - (r_t -\varphi(s_t) + \varphi(s_{t+1})) \right] \\
    &\triangleright\text{By telescoping sums} \\
    &= \min_{\pi : \mathcal{S} \rightarrow \mathcal{A}} \Exp \left[ \varphi(s_0) - \varphi(s_{h_{s \rightarrow s'}(M, \pi)}) + \sum_{t=0}^{h_{s \rightarrow s'}(M, \pi) - 1} r_\text{max} - r_t \right] \\
    &\triangleright\text{By definition of a finite hitting time, $s_{h_{s \rightarrow s'}(M, \pi)} = s'$} \\
    &= \varphi(s) - \varphi(s') + \min_{\pi : \mathcal{S} \rightarrow \mathcal{A}} \Exp \left[ \sum_{t=0}^{h_{s \rightarrow s'}(M, \pi) - 1} r_\text{max} - r_t \right] \\
    &= \varphi(s) - \varphi(s') + c(s, s') \numberthis \label{eq:shaped-cost}
  \end{align*}
  and that the hitting cost-minimizing policy for a state pair is not changed. Therefore,
  \begin{align*}
    \kappa(M^\varphi) &\\
    &\triangleright\text{By definition of MEHC} \\
    &\geq \max \{ c^\varphi(s, s'), c^\varphi(s', s) \} \\
    &\triangleright\text{By (\ref{eq:shaped-cost})} \\
    &= \max \{ c(s, s') + \varphi(s) - \varphi(s'), c(s', s) + \varphi(s') - \varphi(s) \} \\
    &\triangleright\text{The maximum is no smaller than half of the sum} \\
    &\geq \frac{1}{2} [c(s, s') + c(s', s)] \\
    &\triangleright\text{By (\ref{eq:max-pair})} \\
    &\geq \frac{1}{2} \kappa(M).
  \end{align*}
  We obtain the other half of the inequality by observing $M = (M^\varphi)^{-\varphi}$.
\end{proof}

\section{Details of the numerical experiment in Section~\ref{sec:reset-efficient-rl}}
\label{sec:extra-reset-rl}
We constructed an MDP that models racing around a track. There are $\ell$-many positions on the track represented by states $\{s_1, \dots, s_\ell\}$ and at each of them, there are $(k+1)$-many actions including a good action that takes us to the next state wrapping around like in a track with probability of $(1-\delta)$. The good action with probability of $\delta$ and the other $k$ actions lead to a crashed state at that track location. Once crashed, we are stucked at that crashed location. We are rewarded everytime we reach state $s_\ell$ and thus gain is the inverse of the time to finish a lap.
It is clear that a learner ignorant of the transitions cannot recover an inevitable crash. We modify this MDP by adding a reset action to all the states which transition to the state $s_1$.

We used $\ell = 4, k = 2, \delta = 0.2$ in the reported experiments.

\section{Details of the numerical experiment in Section~\ref{sec:multiple-rewards}}
\label{sec:extra-multiple-reward-exp}

\begin{figure}
  \centering
  \begin{tikzpicture}[->, >=stealth', shorten >=1pt, auto, semithick]
    \tikzstyle{action} = [draw=black,fill=none]
        \node[state] at (-2, 0) (s1) {$s_1$};
        \node[state] at (2, 0) (s2) {$s_2$};

        \node[action, above right=of s1] (s1a1) {$a_1$};
        \node[action] at (-4, 0) (s1a2) {$a_2$};

        \node[action, below left=of s2] (s2a1) {$a_1$};
        \node[action] at (4, 0) (s2a2) {$a_2$};

        \path (s1a2) edge [out=135, in=45, looseness=2] node {$\epsilon$} (s2)
                     edge [bend left] node {$1-\epsilon$} (s1);
        \path (s2a2) edge [out=-45, in=-135, looseness=2] node {$\epsilon$} (s1)
                     edge [bend left] node {$1-\epsilon$} (s2);
        \path (s1a1) edge [bend left] node {$\epsilon$} (s1)
                     edge [bend left] node {$1 - \epsilon$} (s2);
        \path (s2a1) edge [bend left] node {$\epsilon$} (s2)
                     edge [bend left] node {$1 - \epsilon$} (s1);

        \path (s1) edge [-, dashed] node [text=red] {$0, 0$} (s1a1);
        \path (s1) edge [-, dashed] node [text=red] {$1, 0$} (s1a2);
        \path (s2) edge [-, dashed] node [text=red] {$0, 0$} (s2a1);
        \path (s2) edge [-, dashed] node [text=red] {$0, 1$} (s2a2);
    \end{tikzpicture}
  \caption{Circular nodes represent states and square nodes represent actions. The solid edges are labeled by the transition probabilities and the dashed edges are labeled by the two kinds of rewards in red font. For our experiment, we set $\epsilon = 0.1$.}
  \label{fig:multi-reward-example}
\end{figure}
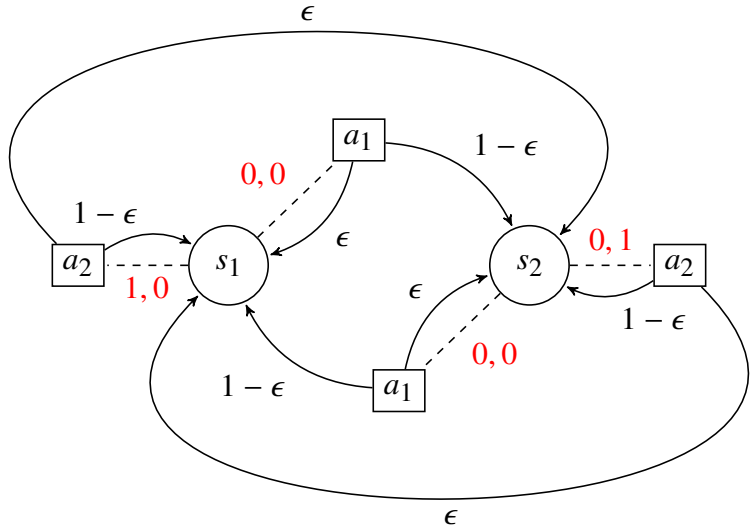

The MDP we used consists of two states with two actions each. We can think of the $a_1$ as the ``switch'' action and $a_2$, the ``stay'' action. However, both actions have a small probability of causing the opposite effect. The two reward functions prefer staying in $s_1$ and $s_2$ respectively, creating a tradeoff. See Figure~\ref{fig:multi-reward-example} for details.

\begin{figure}
  \centering
  \includegraphics[width=0.45\textwidth]{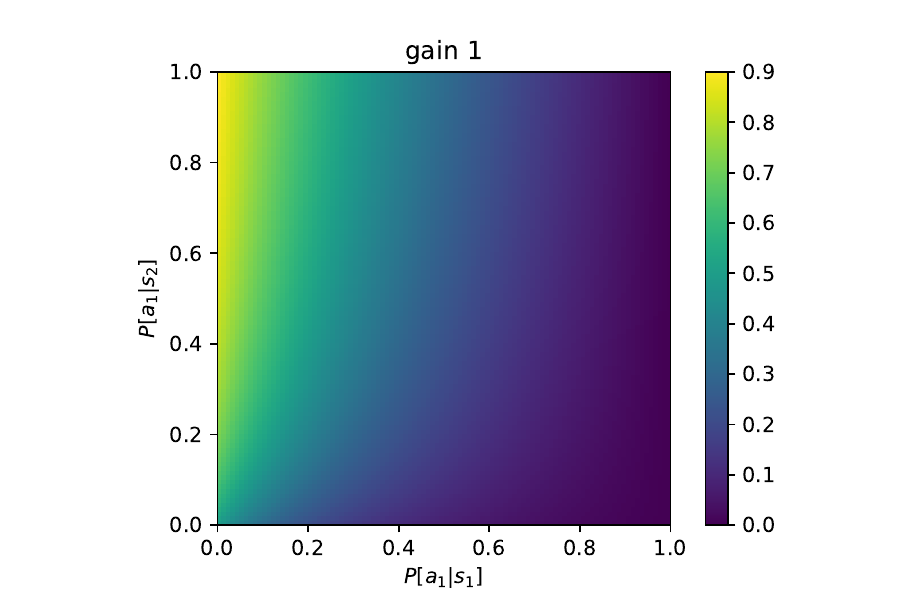}
  \hfill
  \includegraphics[width=0.45\textwidth]{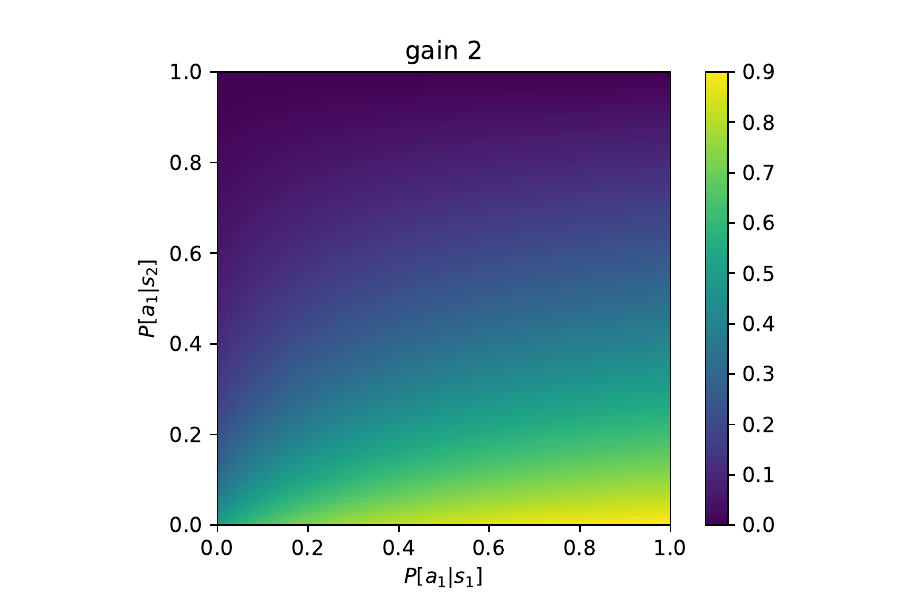}
  \caption{The gains of different policies plotted over the policy space.}
  \label{fig:cramer-gains}
\end{figure}

Since there are only two actions at two states, we can parameterize the Markov stochastic policies by $\rbk*{\pi(a_1|s_1), \pi(a_1|s_2)} \in [0, 1] \times [0, 1]$ and visualize the gains over the policy space (Figure~\ref{fig:cramer-gains}).

\begin{figure}
  \centering
  \includegraphics[width=0.45\textwidth]{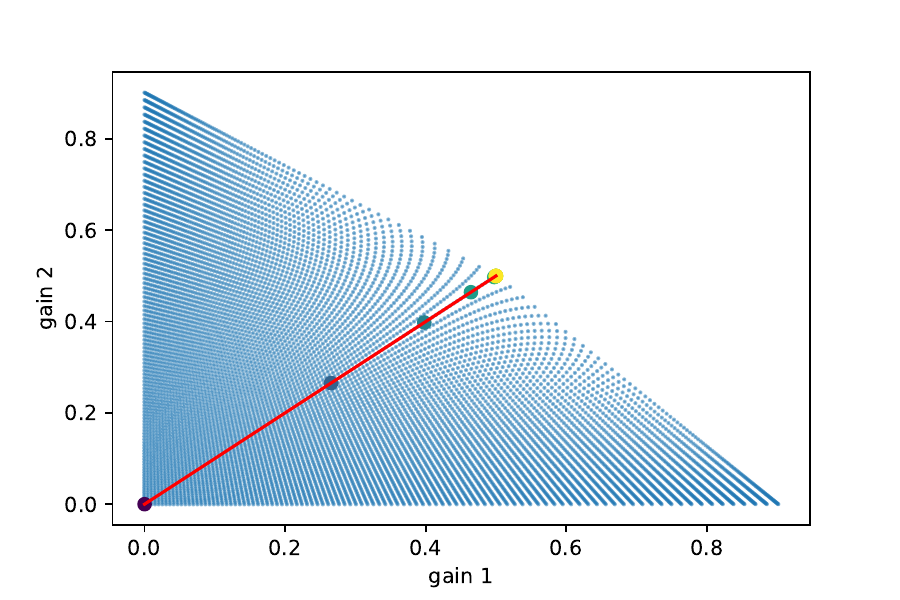}
  \hfill
  \includegraphics[width=0.45\textwidth]{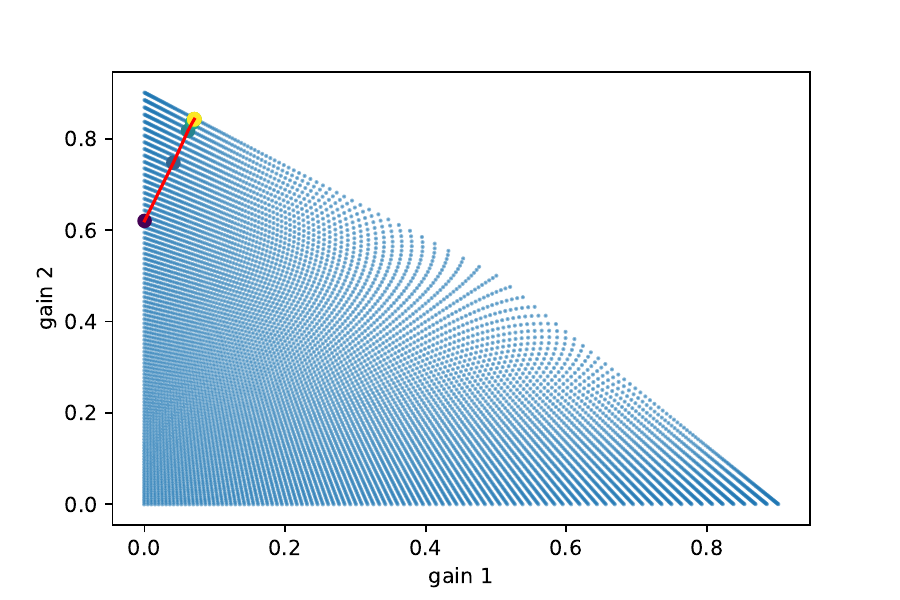}
  \hfill
  \includegraphics[width=0.45\textwidth]{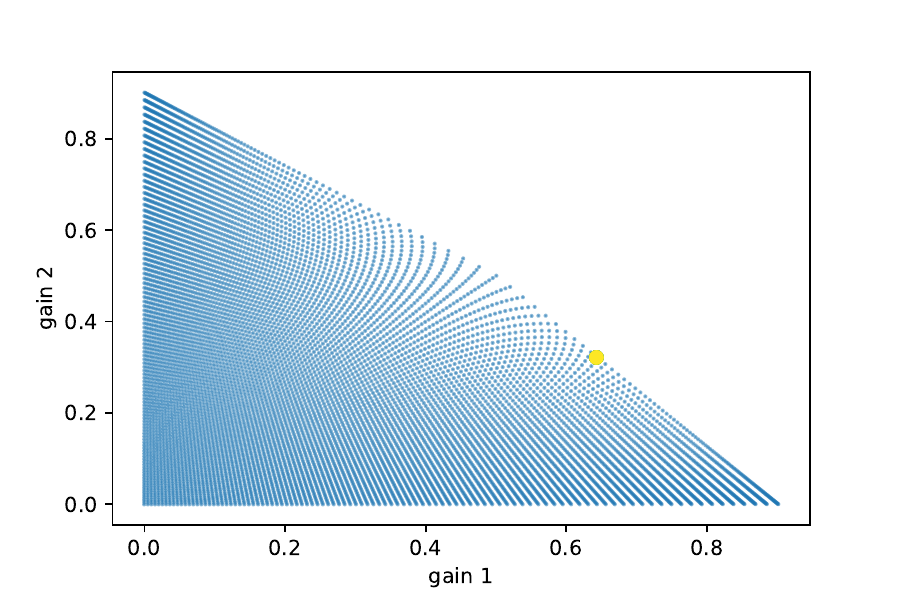}
  \hfill
  \includegraphics[width=0.45\textwidth]{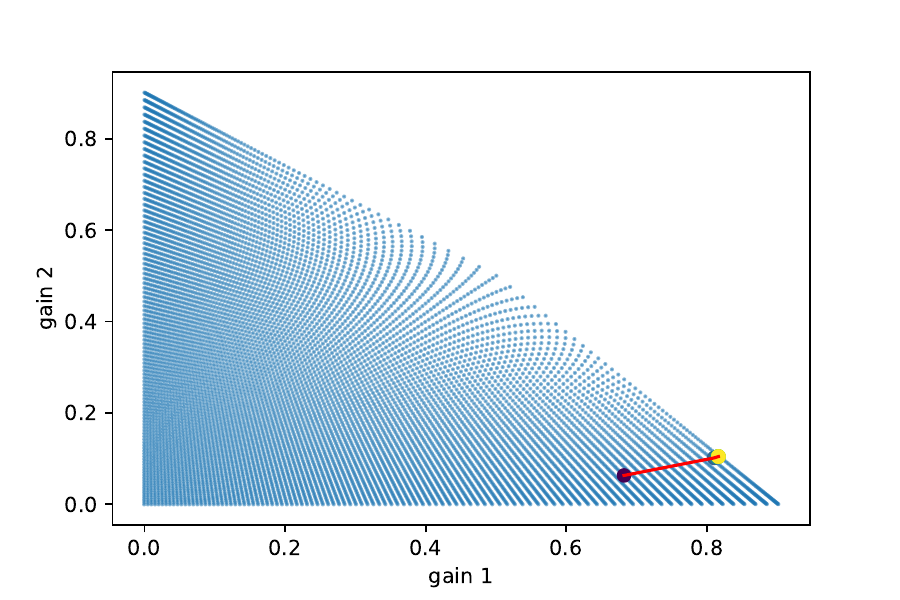}
  \caption{Extra examples of iterates of the proposed Direct-cone policy optimization algorithm color-coded from dark blue to bright yellow. The trajectories have different initial policies. The blue dots are gains of some stochastic policies as seen in Figure~\ref{fig:feasible-gains}.}
  \label{fig:cramer-gains-extra}
\end{figure}

\end{document}